\documentclass[11pt]{article}

\usepackage[top=0.85in,bottom=1in,left=1in,right=1in]{geometry}

\usepackage{amsmath}
\usepackage{amssymb}
\usepackage{array}
\usepackage{tabularx}
\usepackage{booktabs}
\usepackage{graphicx}
\usepackage{float}
\usepackage[section]{placeins}
\usepackage{hyperref}
\usepackage{titling}
\usepackage{textcomp}
\usepackage{adjustbox}
\usepackage{subcaption}

\usepackage[style=apa, backend=biber]{biblatex}
\pretitle{%
  \begin{center}
  \rule{\linewidth}{0.8pt}\par\vspace{0.4em}
  \LARGE\bfseries
}
\posttitle{%
  \par\vspace{0.4em}\rule{\linewidth}{0.8pt}
  \end{center}
}

\preauthor{%
  \begin{center}\normalsize
}
\postauthor{%
  \par\end{center}\vspace{-1.0em}
}

\predate{}
\postdate{}

\title{Lexical Consensus: Grounded Word Learning and Shared Meaning in Artificial Agents}

\author{
P.~M.~Vera\\
Neurocreaciones\\
\texttt{patricio@neurocreaciones.ai}
}

\date{}

\begin{document}

\maketitle
\vspace{-1.0em}

\begin{abstract} Artificial intelligence systems are commonly evaluated through task performance and behavioral imitation, but these evaluations often leave open a deeper question: whether an artificial agent can acquire, stabilize, and use new lexical meanings from grounded experience. This paper introduces \emph{Lexical Consensus}, an experimental framework for studying grounded word learning over a structured perceptual substrate. Using frozen DINOv2 visual embeddings, an artificial lexicon of Carroll-style nonce words, and a family of interpretable lexical learners (centroid, multi-centroid, exemplar \(k\)-NN) plus linear baselines, we test whether agents can acquire artificial labels for visual concepts, generalize them bidirectionally, and stabilize them across controlled acquisition settings. The central empirical finding is a robust, monotonic perceptual-coherence gradient: native categories are easiest to learn, coherent overextensions remain learnable, mid-range disjunctive concepts degrade, and far-disjunctive concepts approach chance. A pre-registered dissociation experiment over CIFAR-100 confirms that this gradient is governed by perceptual distance rather than semantic relatedness: in concept pairs where perceptual and semantic coherence disagree, acquisition accuracy tracks the perceptual predictor (partial \(R^2 = 0.245\), \(p < 10^{-7}\)), while the semantic predictor adds no significant explanatory power (partial \(R^2 = 0.002\), \(p = 0.660\)). Bidirectional evaluation further reveals that exemplar-based mechanisms outperform centroid prototypes in label-to-image retrieval, particularly for coherent but multimodal concepts, exposing a memory-fidelity dimension distinct from naming accuracy. Falsification controls, homogeneous candidate-pool evaluations, and null results on representational restructuring indicate that frozen perceptual geometry both enables lexical grounding and imposes a boundary on what can be acquired without representational adaptation. \end{abstract}

\vspace{-0.6em}


\section{Introduction}

Modern artificial intelligence evaluation is still largely organized around task performance, benchmark accuracy, and behavioral imitation. In natural language processing, this tendency is particularly visible: systems are often evaluated by their ability to produce fluent, correct, or human-like outputs under predefined tasks. These evaluations have been valuable for tracking progress, but they leave open a deeper question: can an artificial agent acquire new lexical meanings from grounded experience, maintain those meanings across unseen instances, and use them bidirectionally as both names and retrieval cues?

This question motivates the present work. Language acquisition has been proposed as a core dimension for future AI evaluation, shifting the emphasis from whether machines can imitate linguistic behavior to whether they can acquire, use, and stabilize language-like mappings in an environment \parencite{vera20xxrethinking}. The first test proposed in that framework asks whether a machine, through direct verbal instruction, can acquire vocabulary for its surroundings without relying only on preloaded labels or task-specific definitions. This paper presents a constrained empirical implementation of that test.

We introduce \emph{Lexical Consensus}, an experimental framework for studying grounded word learning over a structured perceptual substrate. The framework asks whether agents can learn novel word--concept mappings from limited visually grounded examples, whether those mappings generalize bidirectionally from image to label and from label to image, and how their success depends on the relationship between the taught concept and the perceptual geometry available to the agent. The central question is therefore not whether agents can assign arbitrary strings to arbitrary mathematical subsets of images, but which kinds of lexical categories become learnable when instruction operates over an already structured perceptual field.

Instead of using ordinary category names such as ``frog'', ``horse'', or ``ship'', the experiments use artificial lexical items drawn from Lewis Carroll's vocabulary, such as \emph{slithy}, \emph{mimsy}, and \emph{vorpal} \parencite{carroll1871lookingglass}. These labels reduce direct semantic leakage at the lexical level and force meaning to emerge from the experimental association between perception, instruction, and use. Images are encoded using frozen DINOv2 visual embeddings \parencite{oquab2023dinov2}, while learner agents acquire label-specific representations over the resulting embedding space.

The use of a frozen perceptual encoder is central to the design. A pretrained visual model already contains a non-random organization of visual experience. We do not treat this structure as a confound to be hidden, but as the substrate over which acquisition can be measured. Human word learning also begins from structured perception rather than from an unorganized sensory space: broad coherent regions are often named before finer distinctions are learned. In this sense, the relevant question is not whether the perceptual substrate contains structure, but how lexical learning behaves as taught categories become more or less aligned with that structure.

This perspective leads to a concept-carving evaluation. We distinguish native concepts, where one artificial label corresponds to one native visual category; coherent overextensions, where a label groups nearby or perceptually related categories; mid-range disjunctive concepts, where the grouped categories are less coherent; and far-disjunctive concepts, where a label joins distant regions of the perceptual space. This design directly addresses a key concern for grounded word learning over frozen representations: whether apparent acquisition is merely the relabeling of clusters already present in the encoder, or whether the framework can measure how acquisition succeeds and fails as concepts depart from native category boundaries.

Across the experiments, we evaluate four progressively stronger questions. First, can a single agent acquire artificial labels from limited examples and use them bidirectionally, from image to label and from label to image? Second, does acquisition follow a measurable gradient as taught concepts move from native categories to coherent overextensions and then to far-disjunctive concepts? Third, does the effect survive falsification controls, including randomized labels, randomized embeddings, permuted bindings, out-of-vocabulary rejection, harder visual categories, and homogeneous candidate-pool evaluations? Fourth, can multiple agents trained on disjoint seed sets converge toward shared lexical usage, and does consensus reshape internal representations or merely coordinate decisions over a shared perceptual geometry?

The results show that lexical acquisition over frozen perception is not arbitrary set learning. Naming accuracy follows a perceptual-coherence gradient: native categories are easiest to acquire, coherent overextensions remain highly learnable, mid-range disjunctive concepts degrade, and far-disjunctive concepts approach chance. This pattern supports the view that early lexical acquisition is constrained by perceptual coherence. Agents learn labels more reliably when taught concepts correspond to coherent regions of the perceptual space, and performance degrades as the taught concept cuts across distant regions of that space.

Bidirectional evaluation further shows that image-to-label naming and label-to-image retrieval are not redundant. Naming accuracy measures concept--geometry compatibility: whether a new image can be assigned to the correct learned label. Retrieval accuracy measures memory fidelity: whether a learned label can recover valid instances of its extension. In retrieval, exemplar-based mechanisms consistently outperform compressed centroid prototypes, especially for coherent multimodal concepts, while linear discriminative baselines such as logistic regression and linear SVM recover additional structure under hard candidate pools. These comparisons position the centroid learner as an interpretable lexical mechanism rather than a state-of-the-art classifier.

The multi-agent experiments add a second boundary. Agents can converge on a shared Carroll vocabulary, but comparison with a no-feedback baseline and subsequent alignment experiments show that shared perceptual geometry remains the dominant stabilizing force. Consensus refines agreement, but it does not substantially reorganize internal representations under the current architecture. This null result is important: lexical agreement should not be overinterpreted as evidence that language has reshaped perception. Instead, the experiments show how far lexical coordination can go when the perceptual substrate remains frozen.

The contributions of this paper are sixfold: \begin{enumerate} \item We present \emph{Lexical Consensus}, a reproducible experimental framework for evaluating grounded word learning, bidirectional grounding, and lexical stabilization in artificial agents. \item We implement a constrained empirical version of the first language-acquisition test proposed in \textcite{vera20xxrethinking}, connecting acquisition-based AI evaluation to a measurable protocol. \item We show that agents acquire labels according to a perceptual-coherence gradient: native categories are easiest, coherent overextensions remain highly learnable, and arbitrary far-disjunctive concepts degrade toward chance. \item We confirm through a pre-registered dissociation experiment that this gradient is governed by the geometry of the perceptual substrate, not by semantic relatedness or by the metric used to define concept tiers. In pairs where perceptual and semantic distance disagree, acquisition accuracy tracks perceptual distance. \item We show that bidirectional grounding is not redundant: image-to-label naming measures concept--geometry compatibility, while label-to-image retrieval exposes memory fidelity and benefits from exemplar-based mechanisms. \item We report falsification controls, baseline comparisons, homogeneous retrieval-pool evaluations, and null results showing that frozen perceptual geometry both enables lexical acquisition and imposes a boundary on what can be acquired without representational adaptation. \end{enumerate}

Taken together, these results suggest that language acquisition can be studied as an experimentally measurable property of artificial agents, even in a constrained setting. The present work should not be read as a full solution to artificial language acquisition. It is instead a first empirical step toward the broader evaluation program proposed in \textcite{vera20xxrethinking}: moving from imitation-based assessment toward tests that measure how agents acquire, retrieve, stabilize, and share meaning under perceptual constraints.


\section{Experimental Framework}
\label{sec:experimental-framework}

The purpose of the experimental framework is to operationalize grounded lexical acquisition as a measurable process. Following the acquisition-based evaluation program proposed in \textcite{vera20xxrethinking}, we do not evaluate whether an agent already knows a word, but whether an agent can make a novel word become usable for perception, generalization, retrieval, and social coordination. In this setting, a word becomes experimentally meaningful when it can be associated with grounded examples, generalized to unseen instances, used bidirectionally, and, in the multi-agent condition, stabilized across agents.

The framework consists of four layers: (i) a frozen perceptual encoder, (ii) a learnable lexical layer, (iii) an optional multi-agent consensus mechanism, and (iv) a measurement layer for grounding, retrieval, convergence, information transfer, and representational alignment. This separation is deliberate. The perceptual encoder provides a structured visual geometry; the lexical layer learns how artificial words attach to concepts over that geometry; the consensus layer coordinates lexical usage across agents; and the measurement layer determines whether the resulting mappings are stable, generalizable, falsifiable, and bounded by the perceptual substrate.

A central distinction in the revised framework is between \emph{native categories} and \emph{taught lexical concepts}. A native category is a category already present in the dataset, such as frog, horse, ship, automobile, or truck. A taught concept is the extension assigned to an artificial lexical label during the experiment. A taught concept may correspond to one native category, but it may also group multiple categories into a coherent overextension or a disjunctive concept. This distinction allows us to test whether lexical acquisition merely renames native clusters or whether it follows a measurable gradient as taught concepts become more or less aligned with the frozen perceptual geometry.

\subsection{Perceptual Substrate}
\label{subsec:perceptual-substrate}

Let $\mathcal{X}$ be a set of images and let $\mathcal{Y}$ be a set of native visual categories. Each image $x_i \in \mathcal{X}$ has an underlying native category $y_i \in \mathcal{Y}$ used for evaluation and experimental construction, but native category names are removed from the agent-facing protocol. Each image is encoded using a frozen visual encoder:

\begin{equation}
    z_i = f_{\theta}(x_i), \qquad z_i \in \mathbb{R}^{d},
\end{equation}

where $f_{\theta}$ denotes a pretrained DINOv2 encoder whose parameters remain fixed throughout the experiment \parencite{oquab2023dinov2}. In the implementation, DINOv2-small is used as the perceptual substrate, producing normalized embeddings from the CLS token. Embeddings are normalized before distance computations:

\begin{equation}
    \tilde{z}_i = \frac{z_i}{\|z_i\|_2}.
\end{equation}

The frozen encoder is not intended to model full perceptual learning. Rather, it provides a controlled perceptual substrate on which lexical acquisition can be isolated. The presence of structure in this substrate is not treated as a confound. It is the condition under which acquisition can be measured. Human word learning also begins from structured perception rather than from an unorganized sensory field. The experimental question is therefore how lexical learning behaves as taught concepts become more or less compatible with the available perceptual geometry.

Before lexical acquisition is evaluated, the substrate is tested for category separability. We use silhouette score as a diagnostic:

\begin{equation}
    s(i) = \frac{b(i) - a(i)}{\max\{a(i), b(i)\}},
\end{equation}

where $a(i)$ is the mean intra-cluster distance for image $i$, and $b(i)$ is the mean distance to the nearest alternative cluster. The average silhouette score is:

\begin{equation}
    S = \frac{1}{n}\sum_{i=1}^{n}s(i).
\end{equation}

This score is used only as a substrate diagnostic and concept-selection descriptor, not as evidence of lexical acquisition. The distinction is important because the visual space may already contain category structure before any artificial word is learned.

The primary dataset for the concept-carving experiments is CIFAR-10 \parencite{krizhevsky2009learning}, which contains 10 categories (\emph{airplane}, \emph{automobile}, \emph{bird}, \emph{cat}, \emph{deer}, \emph{dog}, \emph{frog}, \emph{horse}, \emph{ship}, and \emph{truck}), each with 5,000 training images and 1,000 test images at \(32 \times 32\) pixels. The dissociation experiment (Section~\ref{sec:exp008_dissociation}) extends to CIFAR-100, which provides 100 fine-grained classes organized into 20 superclasses, with 500 training images and 100 test images per class. All images are encoded once using the frozen DINOv2-small encoder and cached; no data augmentation is applied.

\subsection{Artificial Lexicon}
\label{subsec:artificial-lexicon}

The experimental lexicon is a set of artificial labels:

\begin{equation}
    \mathcal{L} = \{\ell_1, \ell_2, \ldots, \ell_k\}.
\end{equation}

The labels are drawn from Lewis Carroll's invented vocabulary, including items such as \emph{slithy}, \emph{mimsy}, and \emph{vorpal} \parencite{carroll1871lookingglass}. These labels are phonotactically plausible but experimentally ungrounded: they are not presented to the agents as definitions, category names, or semantic descriptions. Their meaning must be inferred from association with visual examples.
Because no component of the architecture processes the label strings
linguistically---labels enter the system only as opaque identifiers, and no
text encoder is ever applied to them---the specific choice of Carroll
vocabulary cannot leak semantic information into acquisition. Any phonotactic
or literary resonance of the labels is, by construction, invisible to the
learners.

Let $\mathcal{C}$ denote the set of taught lexical concepts. Each concept $c_j \in \mathcal{C}$ is defined as a subset of native categories:

\begin{equation}
    c_j \subseteq \mathcal{Y}.
\end{equation}

A native concept contains exactly one category, while a disjunctive or overextended concept contains more than one category. The extension of a concept is the set of images whose native category belongs to that concept:

\begin{equation}
    \mathcal{X}_{c_j}
    =
    \{x_i \in \mathcal{X}: y_i \in c_j\}.
\end{equation}

The hidden experimental mapping between artificial labels and taught concepts is:

\begin{equation}
    h: \mathcal{L} \rightarrow \mathcal{C}.
\end{equation}

The mapping $h$ is used only for evaluation and data construction. The agent observes paired examples of the form $(x_i,\ell_j)$ and must infer how each artificial label attaches to a region, union of regions, or extension in the embedding space.

\subsection{Concept Types and Perceptual Coherence}
\label{subsec:concept-types}

To test whether lexical acquisition extends beyond relabeling native visual categories, we introduce a concept-carving evaluation. Instead of considering only mappings where one artificial label corresponds to one native category, we define taught concepts with varying degrees of alignment to the frozen perceptual geometry.

We distinguish four concept tiers:

\begin{enumerate}
    \item \textbf{Native concepts}: each lexical label corresponds to exactly one native visual category.
    \item \textbf{Near-disjunctive concepts}: each label groups two perceptually or semantically coherent categories, approximating early overextension such as using one broad word for related animals or vehicles.
    \item \textbf{Mid-disjunctive concepts}: each label groups categories with intermediate perceptual distance.
    \item \textbf{Far-disjunctive concepts}: each label groups perceptually distant categories, producing arbitrary unions that cut across the dominant geometry of the encoder.
\end{enumerate}

The motivation is developmental rather than purely set-theoretic. Early word learning does not typically begin by assigning labels to arbitrary subsets of the world. Children often stabilize broad but coherent categories before refining them into narrower distinctions. The concept-carving evaluation therefore tests whether a lexical learner succeeds when the taught concept preserves perceptual coherence and fails when the taught concept violates it.

For each native category $y \in \mathcal{Y}$, we define its centroid in the frozen embedding space:

\begin{equation}
    \bar{z}_{y}
    =
    \frac{1}{|\mathcal{X}_{y}|}
    \sum_{x_i \in \mathcal{X}_{y}}
    \tilde{z}_i,
\end{equation}

with normalized form

\begin{equation}
    \tilde{\bar{z}}_{y}
    =
    \frac{\bar{z}_{y}}{\|\bar{z}_{y}\|_2}.
\end{equation}

For a two-member disjunctive concept $c = \{y_u,y_v\}$, we compute inter-member distance as:

\begin{equation}
    d_{\mathrm{member}}(c)
    =
    1 -
    \operatorname{sim}
    \left(
    \tilde{\bar{z}}_{y_u},
    \tilde{\bar{z}}_{y_v}
    \right),
\end{equation}

where cosine similarity is:

\begin{equation}
    \operatorname{sim}(u,v) = u^\top v.
\end{equation}

Lower inter-member distance indicates that the members of the taught concept occupy nearby perceptual regions. Higher inter-member distance indicates that the concept joins distant regions. Concept tiers are organized using inter-member distance and native separability diagnostics. An initially explored NMI-based gate was found to be non-discriminative for balanced two-category unions and is therefore not used as evidence for concept selection.

\subsection{Single-Agent Lexical Learners}
\label{subsec:single-agent-learners}

A learner agent $a$ receives a support set for each label:

\begin{equation}
    \mathcal{S}_{a,\ell}
    =
    \{(x_i,\ell): x_i \in \mathcal{X}_{h(\ell)}\}.
\end{equation}

The number of support examples per label is denoted by $m_{\ell}$. The primary interpretable learner is a centroid-based lexical learner. For each label, the agent constructs a centroid in embedding space:

\begin{equation}
    c_{a,\ell}
    =
    \frac{1}{|\mathcal{S}_{a,\ell}|}
    \sum_{(x_i,\ell) \in \mathcal{S}_{a,\ell}}
    \tilde{z}_i,
\end{equation}

with normalized form:

\begin{equation}
    \tilde{c}_{a,\ell}
    =
    \frac{c_{a,\ell}}{\|c_{a,\ell}\|_2}.
\end{equation}

Given a held-out image $x$, the centroid learner assigns the label whose centroid has maximal cosine similarity:

\begin{equation}
    \hat{\ell}_a(x)
    =
    \arg\max_{\ell \in \mathcal{L}}
    \operatorname{sim}(\tilde{z}, \tilde{c}_{a,\ell}).
\end{equation}

The centroid learner is not proposed as a state-of-the-art classifier. It is used as an interpretable lexical mechanism: a compressed prototype of the examples associated with a word. To contextualize this mechanism, the experiments also include additional learners and baselines:

\begin{enumerate}
    \item \textbf{Centroid}: one normalized prototype per label.
    \item \textbf{Multi-centroid}: multiple sub-centroids per label, allowing a concept to retain more than one anchor.
    \item \textbf{Exemplar k-NN}: an episodic learner that retains support examples and scores candidates by similarity to stored instances.
    \item \textbf{Logistic regression}: a linear discriminative probe over frozen embeddings.
    \item \textbf{Linear SVM}: a second linear discriminative baseline over the same embeddings.
    \item \textbf{Random}: a chance-level baseline.
\end{enumerate}

All learners operate over the same frozen embeddings and the same paired episodes. This makes it possible to distinguish the role of the perceptual substrate from the role of the lexical mechanism.

For notation, we define a general learner-specific score function:

\begin{equation}
    s_a(x,\ell) \in \mathbb{R},
\end{equation}

where higher values indicate stronger association between image $x$ and label $\ell$. For centroid learners, $s_a(x,\ell)=\operatorname{sim}(\tilde{z},\tilde{c}_{a,\ell})$. For exemplar, multi-centroid, and linear learners, $s_a$ is defined by the corresponding learner's scoring rule.

The centroid learner is structurally equivalent to a prototypical network
\parencite{snell2017prototypical} with a frozen encoder and cosine distance,
but without a learned metric space. We omit learned-metric variants by design,
as the goal is to isolate lexical acquisition from metric learning. The
logistic regression and linear SVM baselines partially address the capacity gap
by providing linear discriminative boundaries over the same frozen embeddings.

\subsection{Bidirectional Grounding}
\label{subsec:bidirectional-grounding}

The framework evaluates lexical acquisition in two directions. Condition 1 evaluates image-to-label naming. Given a held-out test set $\mathcal{T}$, the agent predicts:

\begin{equation}
    \hat{\ell}_a(x)
    =
    \arg\max_{\ell \in \mathcal{L}}
    s_a(x,\ell).
\end{equation}

The naming accuracy is:

\begin{equation}
    \operatorname{Acc}_{C1}
    =
    \frac{1}{|\mathcal{T}|}
    \sum_{(x_i,\ell_i) \in \mathcal{T}}
    \mathbb{I}\left[\hat{\ell}_a(x_i) = \ell_i\right].
\end{equation}

Condition 2 evaluates the inverse direction: label-to-image retrieval. Given a label $\ell$ and a candidate pool $\mathcal{P} \subset \mathcal{X}$, the agent selects the candidate image with the highest association to the label:

\begin{equation}
    \hat{x}_a(\ell)
    =
    \arg\max_{x \in \mathcal{P}}
    s_a(x,\ell).
\end{equation}

The retrieval accuracy is:

\begin{equation}
    \operatorname{Acc}_{C2}
    =
    \frac{1}{|\mathcal{Q}|}
    \sum_{(\ell_j,\mathcal{P}_j) \in \mathcal{Q}}
    \mathbb{I}
    \left[
    \hat{x}_a(\ell_j) \in \mathcal{X}_{h(\ell_j)}
    \right],
\end{equation}

where $\mathcal{Q}$ is a set of label-to-image retrieval queries. This condition prevents the experiment from being reduced to one-way classification. A label may be useful for naming while still failing to retrieve valid instances, or may retrieve valid instances while remaining difficult to discriminate under competing labels. C1 and C2 therefore expose distinct dimensions of grounded word learning.

To measure the stability of a naming decision, we compute a learner-specific margin:

\begin{equation}
    M_a(x)
    =
    s_a(x,\hat{\ell})
    -
    \max_{\ell' \neq \hat{\ell}}
    s_a(x,\ell').
\end{equation}

Margins are interpreted within a learner, not across learners, because score scales differ between cosine similarity, exemplar voting, and linear decision functions.

\subsection{Candidate Pools for Retrieval}
\label{subsec:candidate-pools}

Retrieval performance depends on the composition of the candidate pool. To avoid confounding retrieval accuracy with pool construction, we evaluate C2 under homogeneous candidate-pool conditions. Each pool contains at least one valid image from the target concept and a controlled set of distractors. The pool types include:

\begin{enumerate}
    \item \textbf{Standard N-way}: candidates are sampled from the concepts present in the episode.
    \item \textbf{Small random pool}: five candidates.
    \item \textbf{Medium random pool}: ten candidates.
    \item \textbf{Large random pool}: twenty-five candidates.
    \item \textbf{Hard nearest-neighbor pool}: distractors are selected from visually close non-target examples in embedding space.
    \item \textbf{OOV-25}: approximately one quarter of the candidate pool consists of out-of-vocabulary distractors.
    \item \textbf{OOV-50}: approximately one half of the candidate pool consists of out-of-vocabulary distractors.
\end{enumerate}

These pool conditions are used to test whether retrieval results are robust to candidate-set size, hard negatives, and out-of-vocabulary contamination.

\subsection{Multi-Agent Consensus}
\label{subsec:multi-agent-consensus}

Let $\mathcal{A} = \{a_1, a_2, \ldots, a_N\}$ be a population of learner agents. Each agent receives a different seed subset, producing potentially different lexical representations for the same label. For an image $x$ at round $r$, each agent emits a label assignment:

\begin{equation}
    \hat{\ell}_{a,r}(x)
    =
    \arg\max_{\ell \in \mathcal{L}}
    s_{a,r}(x,\ell).
\end{equation}

The empirical label distribution across agents is:

\begin{equation}
    p_r(\ell \mid x)
    =
    \frac{1}{N}
    \sum_{a \in \mathcal{A}}
    \mathbb{I}\left[\hat{\ell}_{a,r}(x) = \ell\right].
\end{equation}

A consensus label is accepted when the maximum agreement exceeds a threshold $\tau$:

\begin{equation}
    \ell_r^{*}(x)
    =
    \arg\max_{\ell \in \mathcal{L}} p_r(\ell \mid x),
    \qquad
    \max_{\ell \in \mathcal{L}} p_r(\ell \mid x) \geq \tau.
\end{equation}

In the current implementation, $\tau = 0.70$. The per-round agreement score is:

\begin{equation}
    A_r =
    \frac{1}{|\mathcal{T}|}
    \sum_{x \in \mathcal{T}}
    \max_{\ell \in \mathcal{L}} p_r(\ell \mid x).
\end{equation}

Unanimity is measured as:

\begin{equation}
    U_r =
    \frac{1}{|\mathcal{T}|}
    \sum_{x \in \mathcal{T}}
    \mathbb{I}
    \left[
    \max_{\ell \in \mathcal{L}} p_r(\ell \mid x) = 1
    \right].
\end{equation}

Consensus accuracy is computed by comparing the accepted consensus label with the hidden evaluation mapping:

\begin{equation}
    \operatorname{Acc}^{cons}_r =
    \frac{1}{|\mathcal{T}_r^{cons}|}
    \sum_{x_i \in \mathcal{T}_r^{cons}}
    \mathbb{I}
    \left[
    \ell_r^{*}(x_i) = \ell_i
    \right],
\end{equation}

where $\mathcal{T}_r^{cons}$ is the subset of test images for which the consensus threshold is reached at round $r$.

\subsection{Information-Theoretic Measures}
\label{subsec:information-theoretic-measures}

The consensus process can be described as entropy reduction in label space. For each image $x$, the entropy of the label distribution at round $r$ is:

\begin{equation}
    H_r(L \mid x)
    =
    - \sum_{\ell \in \mathcal{L}}
    p_r(\ell \mid x)
    \log p_r(\ell \mid x).
\end{equation}

The average conditional entropy across the test set is:

\begin{equation}
    H_r(L \mid X)
    =
    \frac{1}{|\mathcal{T}|}
    \sum_{x \in \mathcal{T}}
    H_r(L \mid x).
\end{equation}

Lower entropy indicates stronger agreement among agents. However, entropy alone does not establish grounding. A population can agree on the wrong label, or collapse into a single label. For that reason, we also compute the mutual information between image identity and assigned lexical labels:

\begin{equation}
    I(X;L) = H(L) - H(L \mid X),
\end{equation}

where $X$ denotes image identity and $L$ denotes the assigned artificial label. In the implementation, $H(L)$ is estimated from the empirical marginal label distribution across images, and $H(L \mid X)$ is estimated as the mean entropy of the per-image label distributions across agents. We report this grounding-level normalized mutual information, denoted
\(\mathrm{NMI}_{G}\), as:
\[
\mathrm{NMI}_{G}(X,L) = \frac{I(X;L)}{H(L)}.
\]
This quantity is distinct from the average-normalized NMI used in the
pre-episode concept-gate diagnostic audited in Appendix~\ref{app:nmi-audit},
which we denote \(\mathrm{NMI}_{\mathrm{gate}}\).

If $H(L)=0$, normalized mutual information is defined as $0$, since this corresponds to lexical collapse rather than successful grounding. This measure captures how informative the assigned labels are with respect to image identity under the observed label distribution.

\subsection{Representational Alignment and Divergence}
\label{subsec:alignment-divergence}

To distinguish lexical agreement from representational change, we track centroid movement and inter-agent alignment over time. For centroid-based agents, the per-label centroid drift between rounds is:

\begin{equation}
    D_{a,\ell}^{(r)}
    =
    1 -
    \operatorname{sim}
    \left(
    \tilde{c}^{(r)}_{a,\ell},
    \tilde{c}^{(r-1)}_{a,\ell}
    \right).
\end{equation}

The cumulative drift for label $\ell$ is:

\begin{equation}
    CD_{a,\ell}^{(R)}
    =
    \sum_{r=1}^{R}
    D_{a,\ell}^{(r)}.
\end{equation}

To measure inter-agent representational alignment, we compute the average pairwise distance between agents' centroids for the same label:

\begin{equation}
    \operatorname{AlignDist}_{\ell}^{(r)}
    =
    \frac{2}{N(N-1)}
    \sum_{i<j}
    \left[
    1 -
    \operatorname{sim}
    \left(
    \tilde{c}^{(r)}_{a_i,\ell},
    \tilde{c}^{(r)}_{a_j,\ell}
    \right)
    \right].
\end{equation}

An alignment gain is then defined as:

\begin{equation}
    G_{\ell}
    =
    \operatorname{AlignDist}_{\ell}^{(0)}
    -
    \operatorname{AlignDist}_{\ell}^{(R)}.
\end{equation}

Positive values of $G_{\ell}$ indicate decreasing inter-agent centroid distance. This metric tests whether shared lexical interaction merely coordinates decisions or also reshapes internal category representations.

To test regional divergence, the agent population can be partitioned into clusters:

\begin{equation}
    \mathcal{A} = \mathcal{A}_1 \cup \mathcal{A}_2 \cup \cdots \cup \mathcal{A}_q.
\end{equation}

For each cluster $k$ and label $\ell$, we define a cluster centroid:

\begin{equation}
    \bar{c}^{(r)}_{k,\ell}
    =
    \frac{1}{|\mathcal{A}_k|}
    \sum_{a \in \mathcal{A}_k}
    \tilde{c}^{(r)}_{a,\ell}.
\end{equation}

The between-cluster lexical divergence for label $\ell$ is:

\begin{equation}
    B_{\ell}^{(r)}
    =
    \frac{2}{q(q-1)}
    \sum_{u<v}
    \left[
    1 -
    \operatorname{sim}
    \left(
    \bar{c}^{(r)}_{u,\ell},
    \bar{c}^{(r)}_{v,\ell}
    \right)
    \right].
\end{equation}

If lexical feedback acts as an independent representational attractor, isolated clusters should drift toward distinct centroids over time. If perception dominates, between-cluster distances should remain stable or decrease despite isolation and transmission noise.

\subsection{Falsification Controls and Baselines}
\label{subsec:falsification-controls-baselines}

A central requirement of the framework is that positive results must survive falsification. The experiments therefore include several controls designed to rule out trivial explanations.

\paragraph{Random-label control.}
Labels are randomly assigned to examples. If an agent still performs well, the experiment would be vulnerable to label-frequency, ordering, or procedural artifacts.

\paragraph{Balanced-label and repeated-scramble controls.}
Balanced-label controls ensure that each artificial label appears with controlled frequency. Repeated scrambles quantify the distribution of performance under randomized assignments, allowing apparent above-chance behavior to be interpreted statistically rather than anecdotally.

\paragraph{Random-embedding control.}
Visual embeddings are replaced by random vectors. This tests whether performance depends on the perceptual substrate. If grounding disappears under random embeddings, the result cannot be explained by the lexical mechanism alone.

\paragraph{Permuted-binding control.}
The correspondence between images and embeddings is permuted. This breaks the binding between perception and label while preserving marginal distributions. Collapse under this condition indicates that correct image--embedding--label binding is essential.

\paragraph{Out-of-vocabulary rejection.}
Agents are tested on examples outside the learned lexical space. A valid grounding system should not confidently force all unknown inputs into known labels. We define the maximum score:

\begin{equation}
    s_{\max}(x) =
    \max_{\ell \in \mathcal{L}}
    s_a(x,\ell).
\end{equation}

An input is rejected as out-of-vocabulary when:

\begin{equation}
    s_{\max}(x) < \gamma,
\end{equation}

where $\gamma$ is a rejection threshold. Performance is evaluated using AUROC over in-vocabulary and out-of-vocabulary examples.

\paragraph{Hard-category and concept-carving controls.}
The framework is tested on visually harder categories and on concept tiers that vary in perceptual coherence. These controls evaluate whether lexical acquisition is restricted to easily separable native categories or whether it degrades in a measurable way as the taught concept moves away from native category structure.

\paragraph{No-feedback baseline.}
In the multi-agent setting, a no-feedback baseline is used to determine whether consensus interaction improves performance or whether agents already agree because they share the same perceptual encoder.

\paragraph{Random-feedback control.}
When mutable lexical adapters are introduced, random feedback provides a control for whether alignment is caused by meaningful consensus or by generic regularization and repeated updates.

\paragraph{Learner baselines.}
Centroid-based acquisition is compared against multi-centroid, exemplar k-NN, logistic regression, linear SVM, and random baselines. These comparisons clarify whether observed effects reflect the interpretability of the lexical mechanism, the geometry of the substrate, or the capacity of stronger discriminative learners.

\subsection{Traceability and Reproducibility}
\label{subsec:traceability-reproducibility}

All experiments are logged as structured events. The ledger records image identifiers, concept assignments, labels, learner type, support/query splits, candidate-pool construction, confidence scores, consensus outcomes, and per-round metrics. In later experiments, these events are also stored in a graph database, allowing the acquisition and consensus process to be queried as a temporal graph.

This traceability layer is part of the experimental design. Since the target phenomenon is not only final accuracy but the acquisition and stabilization of lexical mappings, the path by which labels are acquired must remain inspectable. This supports reproducibility, error analysis, paired statistical testing, and future extensions to larger vocabularies, more agents, richer visual domains, and more complex interaction topologies.

\subsection{Scope of the Framework}
\label{subsec:scope-framework}

The framework should be interpreted as a constrained test of grounded lexical acquisition, not as a full model of human language acquisition. The agents do not acquire perception from scratch, do not learn syntax, and do not interact with an embodied environment. What the framework tests is narrower but experimentally precise: whether artificial labels can become stable, bidirectional, and socially shareable mappings over a perceptual substrate, and how this process changes as taught concepts vary in perceptual coherence.

This scope is important for interpreting the results. A positive result indicates that lexical grounding can be operationalized under controlled conditions. A gradient across concept tiers indicates that acquisition is constrained by perceptual coherence, not arbitrary set membership. A null result in representational alignment indicates that shared words do not necessarily reshape internal representations when perception is frozen. Together, these outcomes define the current boundary of the architecture: agents can acquire and coordinate lexical decisions over shared perception, but frozen perceptual geometry both enables and limits what can be learned.


\section{Results}
\label{sec:results}

We organize the results around six questions: (i) whether the frozen perceptual substrate is structured enough to support lexical learning, (ii) whether agents acquire labels for native visual categories from limited examples, (iii) whether acquisition follows a perceptual-coherence gradient when taught concepts depart from native category boundaries, (iv) whether image-to-label naming and label-to-image retrieval expose distinct capacities, (v) whether the observed effects survive falsification controls and baseline comparisons, and (vi) whether multi-agent consensus reshapes representations or mainly coordinates labels over a shared perceptual substrate.

\subsection{The frozen perceptual substrate provides usable but non-trivial structure}
\label{subsec:results-substrate}

Before testing lexical acquisition, we first evaluated whether the frozen DINOv2-small embedding space provides a viable perceptual substrate. In the initial diagnostic experiment, the categories frog, horse, and ship produced a silhouette score of $0.2826$, exceeding the predefined threshold of $0.25$. The Davies--Bouldin index was $2.2224$. Mean intra-cluster cosine distances were $0.6962$ for frog, $0.5725$ for horse, and $0.6266$ for ship, while inter-cluster distances ranged from $0.8940$ to $0.9008$.

\begin{figure}[t]
    \centering
    \includegraphics[width=0.85\linewidth]{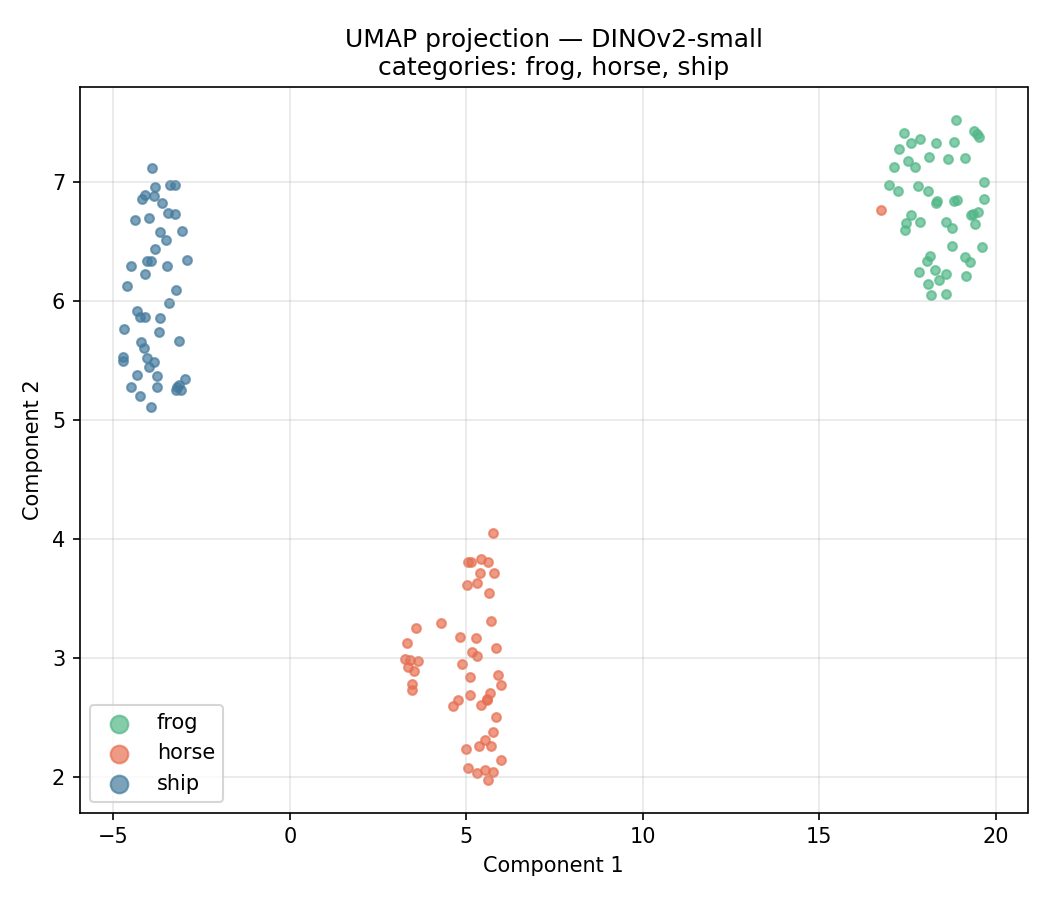}
    \caption{UMAP projection of the frozen DINOv2-small embedding space for the initial visual categories. The substrate is structured before lexical learning, which makes it possible to test how artificial labels attach to that structure.}
    \label{fig:exp000-umap}
\end{figure}

This result is important for interpretation. The embedding space is not random; it already contains visual organization. The experiments therefore do not test whether agents acquire perception from scratch. They test what kinds of lexical meanings can be acquired over a fixed perceptual substrate, and how acquisition succeeds or fails as taught concepts become more or less compatible with that substrate.

\subsection{Native-category labels are acquired from few examples}
\label{subsec:results-native-acquisition}

We first tested whether a single learner agent could acquire arbitrary Carroll labels for native visual categories from small seed sets. The agent was trained with $5$, $10$, or $15$ examples per label and evaluated on $60$ held-out images. Table~\ref{tab:single-agent-native} summarizes the initial result.

\begin{table}[t]
    \centering
    \small
    \begin{tabular}{lccc}
        \toprule
        Seed examples per label & Accuracy & Min label accuracy & Mean margin \\
        \midrule
        5  & 0.9833 & 0.9500 & 0.3766 \\
        10 & 1.0000 & 1.0000 & 0.4045 \\
        15 & 1.0000 & 1.0000 & 0.4097 \\
        \bottomrule
    \end{tabular}
    \caption{Initial native-category Condition 1 naming accuracy across seed sizes. The mapping reaches near-perfect performance with $5$ examples per label and saturates at $10$ examples.}
    \label{tab:single-agent-native}
\end{table}

With $5$ examples per label, the agent reached $0.9833$ accuracy, making one error: a frog instance associated with \emph{slithy} was classified as \emph{vorpal}. With $10$ examples per label, accuracy reached $1.0000$, and the same perfect score was maintained with $15$ examples. This establishes that arbitrary lexical forms can become usable labels for native visual categories under limited instruction.

However, this result alone is not sufficient to establish non-trivial acquisition. A reviewer could reasonably argue that the system is simply assigning new names to clusters already present in the frozen visual encoder. The subsequent concept-carving experiments directly test this concern.

\subsection{Lexical acquisition follows a perceptual-coherence gradient}
\label{subsec:results-concept-gradient}

To determine whether lexical acquisition extends beyond relabeling native clusters, we evaluated taught concepts with different degrees of alignment to the frozen perceptual geometry. We compared four concept tiers: native categories, near-disjunctive coherent overextensions, mid-disjunctive concepts, and far-disjunctive arbitrary unions. The evaluation used homogeneous $3$-way, $5$-shot episodes with $30$ paired episodes and the same learner conditions across tiers.

\begin{figure}[t]
    \centering
    \includegraphics[width=0.92\linewidth]{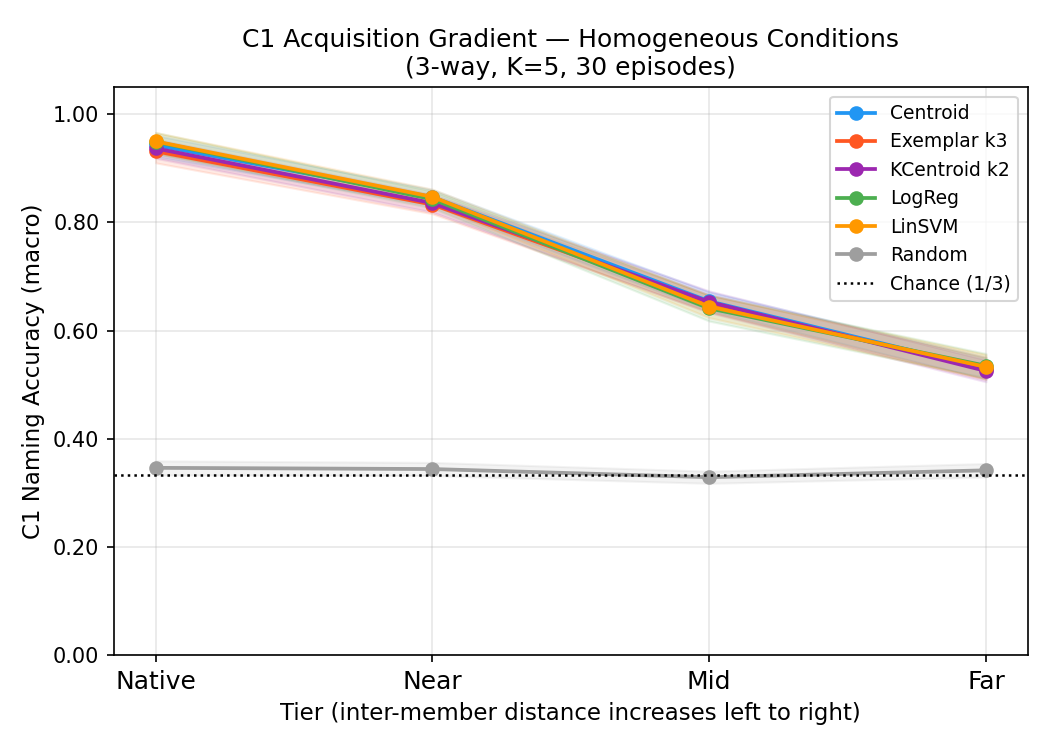}
    \caption{C1 naming accuracy follows a perceptual-coherence gradient. Native categories are easiest, coherent overextensions remain highly learnable, mid-disjunctive concepts degrade, and far-disjunctive concepts approach chance.}
    \label{fig:exp007-c1-gradient}
\end{figure}

Figure~\ref{fig:exp007-c1-gradient} shows the central result: C1 naming accuracy follows a monotonic perceptual-coherence gradient.

Table~\ref{tab:c1-gradient} reports representative $3$-way, $5$-shot C1 accuracies.

\begin{table}[t] \centering \small \begin{tabular}{lcccc} \toprule Concept tier & Centroid [95\% CI] & Exemplar [95\% CI] & LogReg [95\% CI] & Random \\ \midrule Native & 0.943 [0.922, 0.961] & 0.932 [0.910, 0.951] & 0.949 [0.927, 0.966] & 0.346 \\ Near-disjunctive & 0.847 [0.832, 0.861] & 0.833 [0.816, 0.849] & 0.843 [0.827, 0.858] & 0.344 \\ Mid-disjunctive & 0.654 [0.637, 0.672] & 0.650 [0.634, 0.666] & 0.641 [0.618, 0.666] & 0.329 \\ Far-disjunctive & 0.530 [0.510, 0.550] & 0.529 [0.509, 0.551] & 0.535 [0.513, 0.558] & 0.342 \\ \bottomrule \end{tabular} 
\caption{C1 naming accuracy across concept tiers under homogeneous $3$-way, $5$-shot evaluation ($30$ paired episodes, master seed $42$). Values in brackets report 95\% bootstrap confidence intervals. Chance is approximately $0.333$. All non-random learners show the same monotonic degradation as concepts move away from native perceptual coherence.}
 \label{tab:c1-gradient} \end{table}

The gradient is the main empirical result of the paper. Native categories are acquired most reliably. Coherent overextensions, such as grouping nearby or functionally related categories, remain highly learnable. Mid-disjunctive concepts show partial degradation. Far-disjunctive concepts, which join distant regions of the perceptual space, degrade toward chance.

Importantly, the effect is not driven by a particular learner. Centroid, exemplar, and linear learners follow the same C1 ordering across tiers. This indicates that naming accuracy primarily reflects compatibility between the taught concept and the frozen perceptual geometry. Lexical acquisition over frozen perception is therefore not arbitrary set learning; it is constrained by perceptual coherence. The tier differences are statistically robust: bootstrap 95\% confidence intervals do not overlap between adjacent tiers, for example native $[0.922, 0.961]$ versus near-disjunctive $[0.832, 0.861]$, confirming that the C1 gradient reflects genuine differences in concept learnability rather than sampling variation.


\subsection{Perceptual distance governs acquisition independently of semantic relatedness}
\label{sec:exp008_dissociation}

The concept-carving experiments in Section~\ref{subsec:results-concept-gradient}
establish a monotonic gradient: naming accuracy degrades as taught concepts
depart from native perceptual categories. However, the concept tiers were
defined using inter-centroid cosine distance in the same DINOv2 embedding space
over which the learners operate. A reviewer could therefore ask whether the
gradient reflects a genuine constraint from perceptual geometry or an artifact
of defining and measuring concept difficulty with the same metric.

To resolve this ambiguity, we conducted a pre-registered dissociation
experiment; the decision rules, written before data collection, are reproduced
verbatim in Appendix~\ref{app:exp008-preregistration} together with the
registration record.  The experiment introduces an external semantic distance measure and
tests which predictor---perceptual or semantic---governs C1 accuracy in cases
where the two disagree.

\paragraph{Design.}
We extended the evaluation from CIFAR-10 to CIFAR-100, which provides 100
fine-grained classes organized into 20 superclasses \parencite{krizhevsky2009learning}.
For each of the \(\binom{100}{2} = 4{,}950\) class pairs, we computed two
independent distance measures. Perceptual distance was computed as the
inter-centroid cosine distance in the frozen DINOv2-small embedding space,
identical to the metric used in the concept-carving experiments. Semantic
distance was computed as \(1 - \mathrm{Wu\text{-}Palmer}\) similarity over
WordNet synsets, a taxonomic measure derived entirely from lexical ontology and
independent of the visual encoder.

A critical precondition for the dissociation design is that the two predictors
are not strongly correlated. Across all 4,950 pairs, the Pearson correlation
between perceptual and semantic distance was \(r = 0.155\), and the Spearman
correlation was \(\rho = 0.089\). This near-independence provides separation
power: perceptual and semantic distance carry largely non-overlapping
information about concept structure.

Each pair was classified into one of four quadrants based on whether its
perceptual and semantic distances fell above or below their respective medians:
Q1 (both near, \(n = 1{,}255\)), Q2 (both far, \(n = 1{,}444\)), Q3
(perceptually far but semantically near, \(n = 1{,}031\)), and Q4
(perceptually near but semantically far, \(n = 1{,}220\)). Q3 and Q4 are the
dissociation quadrants: pairs where the two criteria disagree. From each
quadrant, 25 pairs were sampled using stratified random sampling
(\(\mathrm{seed}=42\)), yielding 100 pairs total. For each pair, the two member
classes were grouped under a single Carroll label and evaluated in 3-way,
5-shot episodes with 30 paired repetitions, following the same protocol as
Section~\ref{subsec:results-concept-gradient}. In each episode, the union concept competed against two distractor labels assigned to native singleton categories, held fixed across the 30 episodes of that pair and sampled, per the pre-registered specification, from classes outside the pair and outside the CIFAR-100 superclass of either member, to avoid confounding the contrast set with the semantic structure under test. Chance level therefore remains approximately \(0.333\), and episode difficulty is comparable across quadrants.

\begin{figure}[t]
    \centering
    \includegraphics[width=\linewidth]{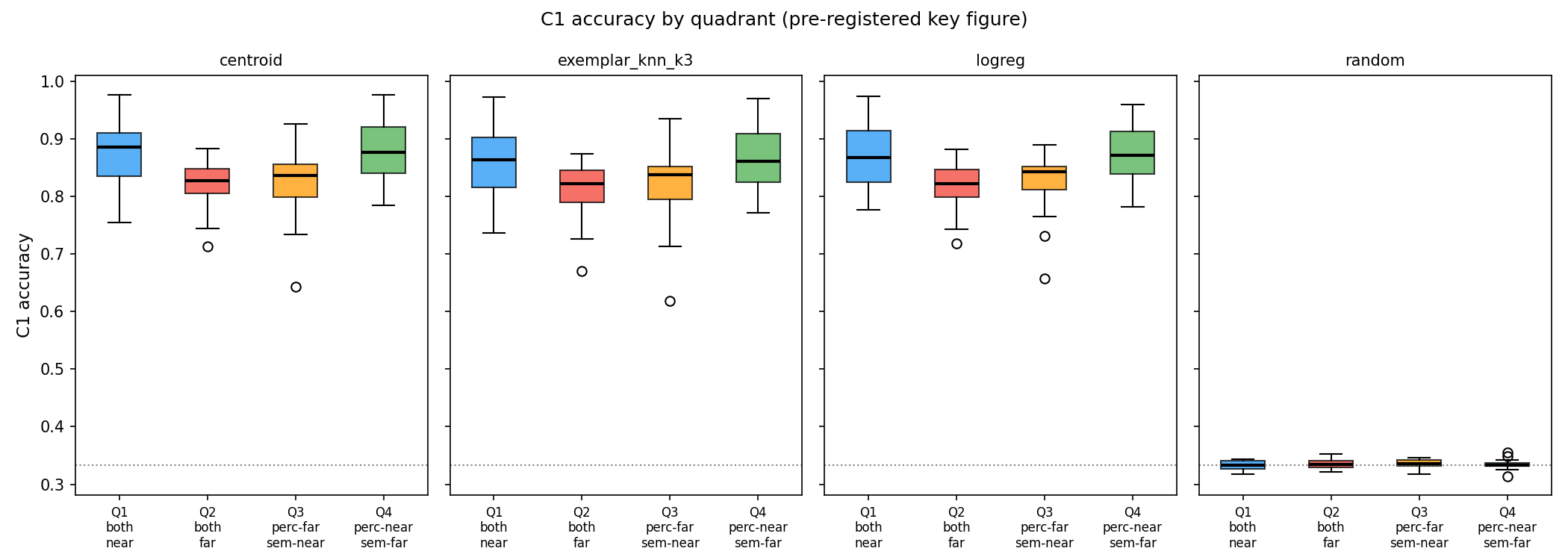}
    \caption{C1 naming accuracy by dissociation quadrant. Pairs are classified
    by whether their perceptual distance (DINOv2 inter-centroid cosine) and
    semantic distance (WordNet Wu-Palmer) fall above or below the median. The
    pattern follows perceptual distance: Q1 (both near) and Q4 (perceptually
    near, semantically far) are comparably easy, while Q2 (both far) and Q3
    (perceptually far, semantically near) are comparably difficult. All
    non-random learners show the same ordering.}
    \label{fig:exp008_quadrant_boxplot}
\end{figure}

\paragraph{Results.}
Figure~\ref{fig:exp008_quadrant_boxplot} shows C1 naming accuracy by quadrant.
The pattern follows perceptual distance, not semantic distance. Q1 (both near)
and Q4 (perceptually near, semantically far) show comparably high C1 accuracy:
\(0.876\;[0.853, 0.899]\) and \(0.877\;[0.857, 0.897]\), respectively. Q2
(both far) and Q3 (perceptually far, semantically near) show comparably lower
accuracy: \(0.822\;[0.805, 0.836]\) and \(0.824\;[0.801, 0.844]\),
respectively. Thus, concept pairs that are perceptually close but semantically
distant are acquired as reliably as concordant-near pairs, while pairs that are
semantically close but perceptually distant degrade. The centroid, exemplar
\(k\)-NN, and logistic regression learners all exhibit the same quadrant
ordering; the random baseline shows no structure.

A multiple regression confirms this pattern quantitatively. With both
predictors in the model, perceptual distance is the dominant predictor of C1
accuracy (\(\beta_{\mathrm{perc}} = -0.212\), \(p < 10^{-7}\); partial
\(R^2 = 0.245\)), while semantic distance contributes no significant additional
explanatory power (\(\beta_{\mathrm{sem}} = -0.013\), \(p = 0.660\); partial
\(R^2 = 0.002\)). A likelihood-ratio test confirms that adding semantic distance
to the perceptual model does not improve fit (\(\chi^2 = 0.20\), \(p = 0.655\)),
whereas adding perceptual distance to the semantic model produces a large
improvement (\(\chi^2 = 28.14\), \(p < 10^{-7}\)). The two predictors exhibit
negligible collinearity (\(\mathrm{VIF}=1.01\)). Full regression coefficients
and the partial-regression visualization are reported in
Appendix~\ref{app:exp008_regression}.

The dissociation subset analysis provides the most direct test. Restricting to
the 50 pairs in Q3 and Q4, where perceptual and semantic distance disagree, the
Spearman correlation between C1 accuracy and perceptual distance is
\(\rho = -0.475\) (\(p = 0.0005\)), strong and in the predicted direction. The
Spearman correlation between C1 accuracy and semantic distance is
\(\rho = +0.327\) (\(p = 0.020\)). This positive association indicates that, in
the dissociation subset, higher semantic distance is associated with higher
acquisition accuracy. The reversal reflects the dominance of the perceptual
predictor: Q4 pairs are easy despite being semantically distant because they
are perceptually near, while Q3 pairs are difficult despite being semantically
near because they are perceptually far.

As a robustness check, we tested whether sharing a CIFAR-100 superclass---a
coarse semantic grouping independent of WordNet---predicts C1 after controlling
for perceptual distance. It does not (\(\beta = 0.015\), \(p = 0.528\)).

The quadrant contrasts, approximately \(0.82\)--\(0.88\), are smaller in
magnitude than the tier gradient reported in
Section~\ref{subsec:results-concept-gradient}, approximately \(0.53\)--\(0.94\).
This difference follows directly from the sampling design rather than from a
weakening of the effect. First, the tier comparison contrasts singleton native
concepts against deliberately constructed far-disjunctive unions targeting the
tail of the distance distribution, whereas all quadrant pairs are two-category
unions whose perceptual distances are determined by a median split
(threshold \(0.712\)) over all \(4{,}950\) pairs: the perceptually near group
(Q1, Q4) has mean inter-centroid distance \(0.586\), and the perceptually far
group (Q2, Q3) has mean inter-centroid distance \(0.784\), a separation of only
\(0.198\). Second, the observed quadrant gap is quantitatively consistent with
the fitted model: the regression slope \(\beta_{\mathrm{perc}} = -0.212\)
predicts a C1 difference of approximately \(0.042\) for this distance
separation, close to the observed difference of \(0.054\). The dissociation
experiment therefore tests the direction and independence of the perceptual
predictor within the central mass of the distance distribution; it is not
designed to reproduce, and should not be expected to reproduce, the full
dynamic range of the tier gradient.

\paragraph{Interpretation.}
This experiment confirms that the perceptual-coherence gradient reported in
Section~\ref{subsec:results-concept-gradient} is not a measurement tautology. The
gradient is not merely an artifact of defining concept difficulty with the same
metric used to evaluate acquisition. Rather, perceptual distance governs
concept learnability even when an independent semantic criterion predicts the
opposite. This supports the central claim of the paper: early lexical
acquisition is constrained by the geometry of the perceptual substrate.


\subsection{Retrieval reveals a distinct memory-fidelity dimension}
\label{subsec:results-c2-retrieval}

Condition 2 evaluates the inverse direction: label-to-image retrieval. This test asks whether a learned label can recover a valid instance of its extension from a candidate pool. Unlike C1, which measures image-to-label naming, C2 measures whether the learned word can function as a retrieval cue.

\begin{figure}[t]
    \centering
    \includegraphics[width=0.92\linewidth]{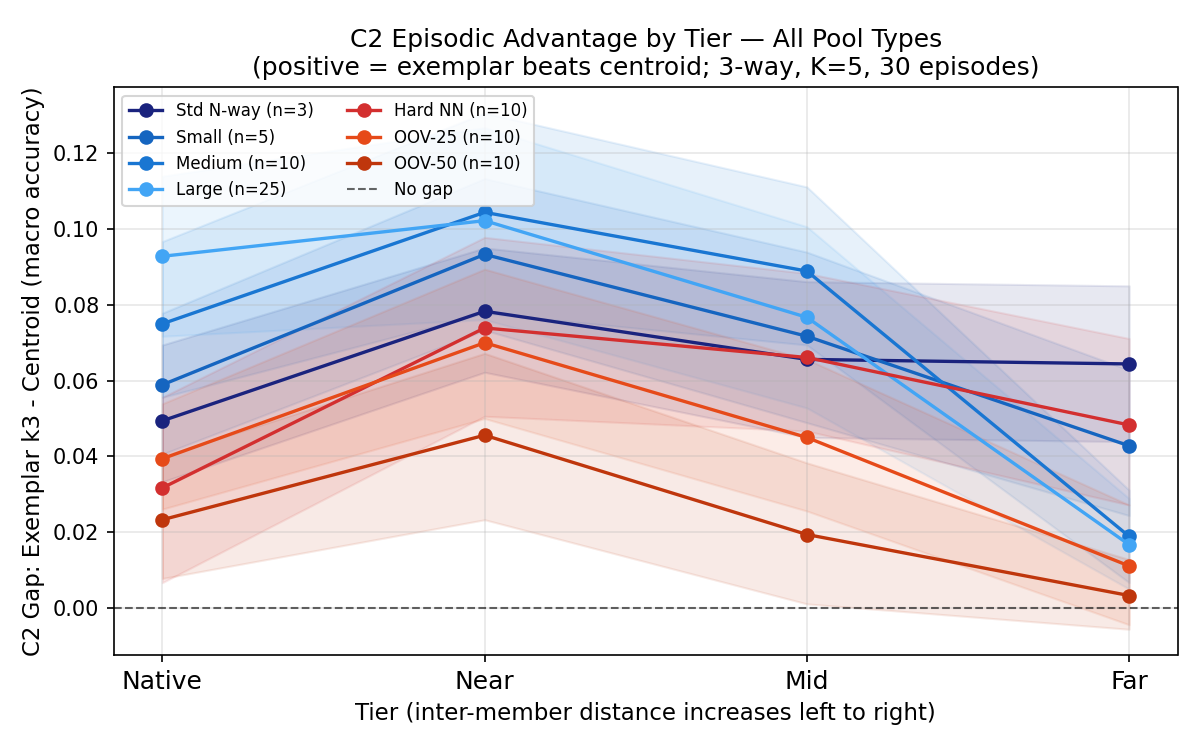}
    \caption{C2 exemplar-over-centroid retrieval gap across concept tiers and candidate-pool constructions. Positive values indicate that exemplar memory retrieves valid instances more accurately than the compressed centroid prototype. The advantage is strongest for coherent overextensions and weaker for arbitrary far-disjunctive concepts.}
    \label{fig:exp007-c2-gap}
\end{figure}

Under homogeneous candidate-pool conditions, exemplar-based retrieval consistently outperforms compressed centroid prototypes. Figure~\ref{fig:exp007-c2-gap} summarizes the exemplar-over-centroid C2 gap across tiers and pool constructions.

Table~\ref{tab:c2-hard-pool} reports the hard-pool C2 results, where distractors are selected as visually close non-target examples.

\begin{table}[t]
    \centering
    \small
    \begin{tabular}{lccccc}
        \toprule
        Concept tier & Centroid [95\% CI] & Exemplar [95\% CI] & LogReg & Gap [95\% CI] & $p$ \\
        \midrule
        Native
        & 0.465 [0.416, 0.514]
        & 0.497 [0.446, 0.548]
        & 0.616
        & +0.032 [0.007, 0.056]
        & 0.017 \\
        Near-disjunctive
        & 0.306 [0.268, 0.342]
        & 0.380 [0.338, 0.419]
        & 0.448
        & +0.074 [0.051, 0.098]
        & $<0.001$ \\
        Mid-disjunctive
        & 0.274 [0.245, 0.305]
        & 0.341 [0.308, 0.374]
        & 0.387
        & +0.066 [0.047, 0.088]
        & $<0.001$ \\
        Far-disjunctive
        & 0.304 [0.274, 0.335]
        & 0.352 [0.319, 0.386]
        & 0.404
        & +0.048 [0.027, 0.071]
        & $<0.001$ \\
        \bottomrule
    \end{tabular}
    \caption{C2 retrieval under hard candidate pools. Exemplar memory improves over compressed prototypes, while logistic regression recovers additional discriminative structure from the frozen
embedding space.}
    \label{tab:c2-hard-pool}
\end{table}

The C2 result should be interpreted differently from the C1 gradient. C1 measures whether the agent can assign the correct label to a new image. C2 measures whether the label can retrieve a valid instance. The exemplar-over-centroid advantage is present across all concept tiers, including native categories (gap $= +0.032$, 95\% CI $[0.007, 0.056]$, paired Wilcoxon $p = 0.017$), confirming that instance-level memory is a generally better retrieval mechanism than a compressed prototype. However, this advantage is not monotonically related to concept distance. It is largest for near-disjunctive concepts ($+0.074$, 95\% CI $[0.051, 0.098]$, $p < 0.001$) and decreases toward far-disjunctive concepts ($+0.048$, 95\% CI $[0.027, 0.071]$, $p < 0.001$). The gap is significantly larger for near-disjunctive than for native concepts ($\Delta_{\mathrm{gap}} = +0.042$, 95\% CI $[0.007, 0.077]$), indicating partial specificity to disjunctive concept structure. The advantage peaks at intermediate perceptual coherence: concepts broad enough for instance fidelity to matter, but still coherent enough for retrieval to remain meaningful.

Linear discriminative baselines, especially logistic regression and linear SVM, often outperform both centroid and exemplar mechanisms under hard retrieval pools. This indicates that the frozen embedding space contains linearly recoverable structure that simple prototype and exemplar mechanisms do not fully exploit. For this reason, we interpret the centroid learner as an interpretable lexical mechanism, not as a state-of-the-art classifier.

\subsection{C2 retrieval is robust to candidate-pool construction}
\label{subsec:results-c2-pools}

\begin{figure*}[t]
    \centering
    \includegraphics[width=0.95\linewidth]{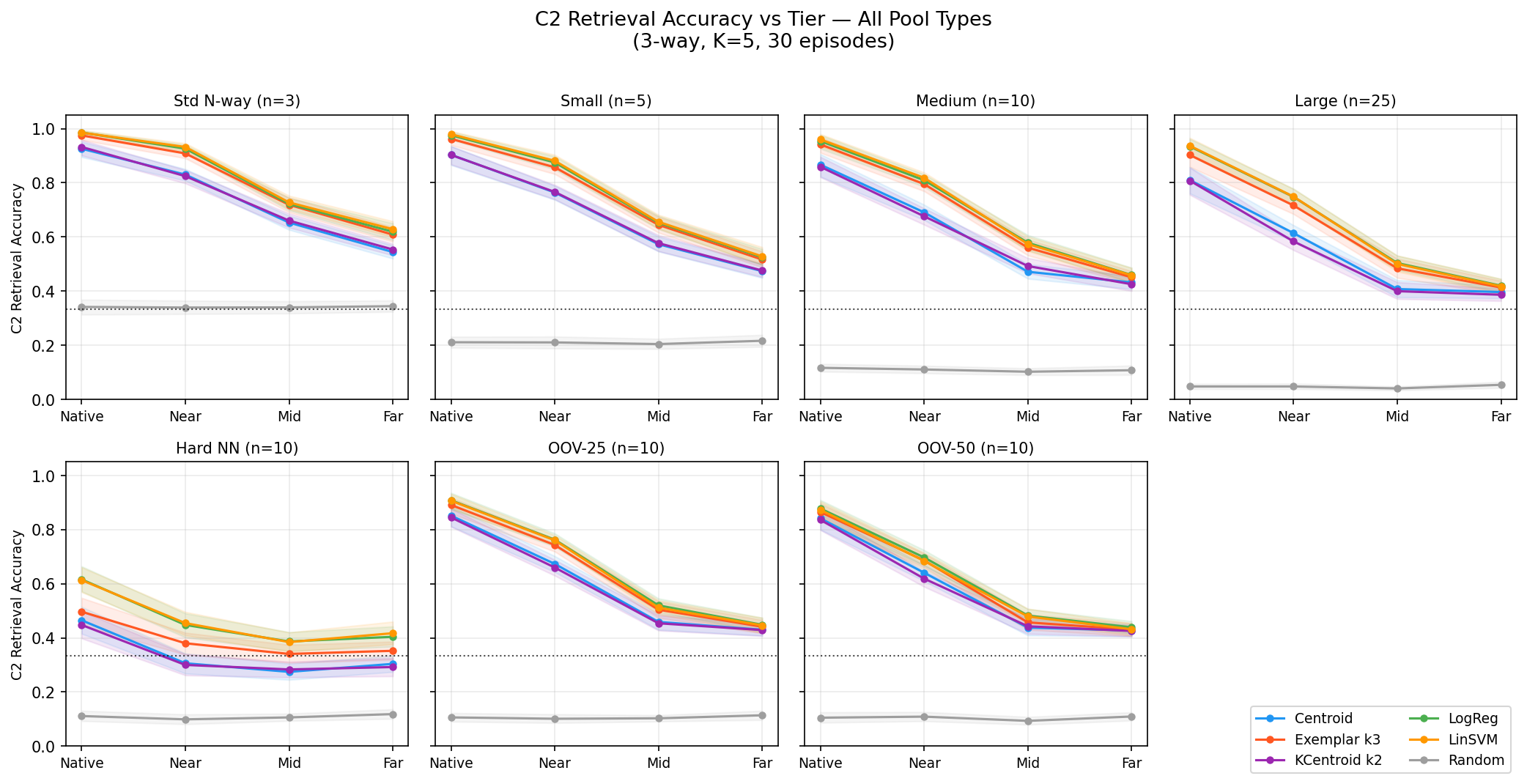}
    \caption{C2 retrieval accuracy across homogeneous candidate-pool constructions. Each pool type is evaluated for native, near-disjunctive, mid-disjunctive, and far-disjunctive concept tiers under the same $3$-way, $5$-shot, $30$-episode setting.}
    \label{fig:exp007-c2-pool-sensitivity}
\end{figure*}

Because retrieval can be sensitive to candidate-pool construction, we evaluated C2 under homogeneous pool conditions: standard N-way pools, small random pools, medium random pools, large random pools, hard nearest-neighbor pools, and OOV-contaminated pools. Figure~\ref{fig:exp007-c2-pool-sensitivity} summarizes the resulting retrieval accuracies.

The exemplar-over-centroid gap remains positive across pool constructions for coherent overextensions. The gap is largest in random pools and smaller under heavy OOV contamination. This confirms that the retrieval effect is not an artifact of a single easy candidate set. At the same time, the results show that retrieval and naming should not be collapsed into one metric. C1 and C2 expose different aspects of grounded lexical acquisition: naming reflects concept--geometry compatibility, while retrieval reflects memory fidelity and candidate-set discrimination.

\subsection{Falsification controls rule out trivial grounding explanations}
\label{subsec:results-controls}

We then tested whether the observed grounding effect could be explained by trivial artifacts such as label frequency, ordering, or the lexical mechanism alone. Table~\ref{tab:controls} summarizes the main grounding controls.

\begin{table*}[t]
    \centering
    \small
    \begin{tabular}{llccc}
        \toprule
        Control & Perturbation & C1 accuracy & C2 accuracy & Interpretation \\
        \midrule
        A & Random labels & 0.5500 & 0.4333 & Partial degradation \\
        B & Random embeddings & 0.3000 & 0.2500 & Near-chance collapse \\
        C & Permuted embeddings & 0.0000 & 0.0000 & Complete collapse \\
        D & OOV rejection & -- & AUROC $=0.9642$ & Strong rejection signal \\
        E & Harder categories & 0.9500 & 0.8667 & Grounding persists \\
        \bottomrule
    \end{tabular}
    \caption{Grounding controls. The effect collapses when perception or image--embedding binding is broken, while harder visual categories remain learnable.}
    \label{tab:controls}
\end{table*}

The random-embedding control reduced C1 accuracy to $0.3000$ and C2 accuracy to $0.2500$, approximately chance-level behavior. The permuted-binding control produced complete collapse in both directions, with \(0.0000\) accuracy for C1 and C2---below chance rather than at chance. This is expected behavior rather than an anomaly: a fixed permutation of the image--embedding binding induces a deterministic relabeling cycle, visible in Figure~\ref{fig:app-exp002-controls}c, where the three labels are systematically exchanged. Errors under this control are structured, not random. These controls show that the lexical layer alone is insufficient: correct grounding depends on the perceptual substrate and on the correct binding between images, embeddings, and labels.

The random-label condition produced a more complex result. Both directions degraded only partially: C2 fell to \(0.4333\) and C1 remained at \(0.5500\), in both cases above their respective chance levels. This motivated the balanced-label and repeated-scramble controls. Under repeated scrambles, mean accuracies returned near chance: \(0.3415\;[95\%~\mathrm{CI}: 0.338, 0.345]\) for C1 and \( 0.3458\;[95\%~\mathrm{CI}: 0.342, 0.349]\) for C2 across 100 scrambles, with no individual scramble exceeding \(0.55\). Thus, random labels can occasionally produce partial apparent structure, but repeated controls show that such behavior is unstable and does not match the robust grounding observed under valid mappings.

The OOV rejection control achieved AUROC $=0.9642$, with an optimal threshold of $0.5289$. The false accept rate was $0.0000$, while the false reject rate was $0.3000$. This suggests that the learned lexical space can support rejection of inputs outside the trained lexical space, although at the cost of rejecting some valid instances.

A diagnostic audit also showed that an initially explored NMI-based concept gate was non-discriminative for balanced two-category unions, producing a constant value under the binary membership formulation. We therefore do not use that value as evidence for concept selection. The final concept tiers are organized using inter-member distance and native separability diagnostics.

\subsection{Multi-agent consensus refines an already strong perceptual baseline}
\label{subsec:results-consensus}

The multi-agent experiments tested whether independently seeded agents could converge on a shared Carroll vocabulary. Agents received disjoint seed sets and interacted through a consensus ledger. We compared a feedback condition against a no-feedback baseline.

\begin{table}[t]
    \centering
    \small
    \begin{tabular}{lccc}
        \toprule
        Condition & Rounds & Final unanimity & Held-out consensus accuracy \\
        \midrule
        Feedback & 6 & 0.9778 & 1.0000 \\
        No feedback & 100 & 0.9333 & 0.9833 \\
        \bottomrule
    \end{tabular}
    \caption{Multi-agent consensus results. Feedback improves unanimity and held-out consensus accuracy, but the no-feedback baseline is already high.}
    \label{tab:multi-agent}
\end{table}

With feedback, held-out consensus accuracy reached $1.0000$ in $6$ rounds, and unanimity rose to $0.9778$. Without feedback, the system remained essentially flat from the first round, with final unanimity $0.9333$ and held-out consensus accuracy $0.9833$. The result is therefore positive but constrained: consensus feedback improves
agreement, but the shared DINOv2 geometry already induces strong alignment
among independently seeded agents. We note explicitly that the no-feedback
baseline operates near ceiling (\(0.9833\) held-out consensus accuracy), which
limits the sensitivity of this design to feedback effects. A more informative
test would seed consensus over mid-disjunctive concepts, where individual
agreement is far from saturation; we identify this as the natural next
experiment (Section~\ref{sec:limitations}).

Information-theoretic measurements reinforce this interpretation. Normalized mutual information reached $0.9659$ from the first round in the no-feedback baseline and $0.9882$ in the feedback condition. Conditional entropy remained low in both cases, indicating that the artificial labels were highly informative with respect to image identity even without feedback. Feedback therefore acts less as the origin of grounding and more as a refinement mechanism over an already aligned perceptual substrate.

\subsection{Consensus does not substantially reshape representations}
\label{subsec:results-representational-null}

We next tested whether lexical consensus reshapes internal category representations. If shared lexical interaction acts as a representational attractor, then inter-agent centroid distances should decrease more under feedback than under the no-feedback baseline.

The result does not support this hypothesis. In the interaction-only reconstruction, mean alignment gain was $0.0016$. In the full operational reconstruction, which includes seed embeddings and accepted interaction examples, mean alignment gain was $0.0015$. The largest gains were small: approximately $0.0038$ for \emph{slithy} and $0.0008$ for \emph{vorpal} in the full reconstruction, with no gain for \emph{mimsy}.

We then introduced a mutable per-agent LexicalAdapter and compared three conditions: frozen adapter, consensus feedback, and random feedback. Consensus feedback did not reduce final projected inter-agent distance relative to the frozen condition, and it performed worse than random feedback. The consensus condition ended with mean projected distance $0.0581$, while the random-feedback condition ended at $0.0393$. Held-out accuracy remained high in all conditions, indicating that lexical performance survived the adapter manipulation, but representational alignment did not become consensus-specific.

Finally, we tested whether agent clusters develop divergent lexical representations under different communication regimes. The Latin regime produced a higher final between-cluster distance ($0.0785$) than the vervet ($0.0051$) and raven ($0.0056$) regimes. However, the mean divergence growth rate in the Latin regime was negative ($-0.00147$), indicating that clusters began farther apart due to disjoint seeding and then slowly converged rather than diverged. Cross-cluster label agreement also remained high at $0.9778$.

Taken together, these results show that lexical agreement should not be overinterpreted as representational transformation. Agents can coordinate labels over a shared perceptual substrate, but the current architecture does not show strong evidence that lexical feedback reorganizes the underlying embedding geometry.

\subsection{Summary of findings}
\label{subsec:results-summary}

The experiments support six main conclusions. First, arbitrary Carroll labels
can be acquired for native visual categories from limited grounded examples.
Second, lexical acquisition over frozen perception follows a
perceptual-coherence gradient: native categories are easiest, coherent
overextensions remain highly learnable, mid-disjunctive concepts degrade, and
far-disjunctive concepts approach chance. Third, a pre-registered dissociation
experiment over CIFAR-100 confirms that this gradient is governed by the
geometry of the perceptual substrate, not by semantic relatedness: perceptual
distance predicts acquisition accuracy (partial \(R^2 = 0.245\),
\(p < 10^{-7}\)), while semantic distance adds no significant explanatory power
after controlling for perception (partial \(R^2 = 0.002\), \(p = 0.660\)).
Fourth, bidirectional grounding is not redundant: C1 naming measures
concept--geometry compatibility, while C2 retrieval measures memory fidelity
and candidate-set discrimination. Fifth, the grounding effect depends on the
visual substrate and correct image--embedding--label binding; it collapses
under random or permuted perceptual conditions and remains robust under harder
categories and homogeneous retrieval-pool evaluations. Sixth, multi-agent
consensus improves lexical agreement but does not substantially reorganize
internal representations.

Thus, \emph{Lexical Consensus} succeeds not as a proof of unrestricted
artificial language acquisition, but as a constrained empirical protocol for
measuring the boundary of grounded lexical learning. The main result is that
early lexical acquisition is not arbitrary set learning. It is the
stabilization of names over perceptually coherent regions, and this constraint
is demonstrably perceptual rather than semantic.


\section{Positioning with Respect to Prior Work}
\label{sec:prior-work-positioning}

The present work is related to several research traditions, but it is not identical to any one of them. Lexical Consensus connects acquisition-based AI evaluation, symbol grounding, grounded word learning, emergent communication, naming games, few-shot learning, and representational alignment. Its contribution is not to introduce the first multi-agent communication setting, nor to claim that arbitrary labels over visual embeddings constitute full language acquisition. Rather, the contribution is a controlled and falsifiable experimental protocol for measuring which lexical concepts become learnable over a structured perceptual substrate, how this learnability changes with perceptual coherence, and how naming, retrieval, consensus, and representation interact under frozen perception.

\subsection{From imitation-based evaluation to acquisition-based evaluation}

The motivation for this paper begins with a limitation in traditional AI evaluation. Since the Turing Test, linguistic behavior has often been evaluated through the lens of imitation: whether a machine can produce outputs that appear human-like to an observer \parencite{turing1950computing}. More recent work has argued for ability-oriented and more general evaluation frameworks, moving beyond static task success toward broader measurements of machine intelligence \parencite{hernandezorallo2017evaluation,chollet2019measure}. The acquisition-based framework proposed in \textcite{vera20xxrethinking} belongs to this broader shift: it argues that language acquisition, rather than imitation alone, should be treated as a core axis for evaluating artificial agents.

Lexical Consensus provides a constrained empirical step in that direction. The aim is not to replace existing benchmarks, but to test a specific capability that ordinary performance benchmarks do not isolate: the ability to acquire a novel lexical mapping from limited grounded examples and use it as both a naming function and a retrieval cue. In this sense, the framework evaluates a learning process rather than only an output state.

The concept-carving experiments extend this acquisition-based view. They show that lexical acquisition over frozen perception is not all-or-nothing. Agents acquire labels most reliably when taught concepts correspond to native perceptual regions, retain high performance for coherent overextensions, and degrade as concepts become arbitrary unions of distant regions. This supports a developmental interpretation: early lexical learning is not arbitrary set learning, but the stabilization of names over perceptually coherent regions before finer distinctions are learned.

\subsection{Symbol grounding and grounded language learning}

The symbol grounding problem asks how symbols can acquire meaning intrinsic to a system rather than remaining dependent on interpretation by an external observer \parencite{harnad1990symbol}. In contemporary NLP, this problem has been revisited in critiques of purely text-based learning, where form alone may be insufficient for meaning \parencite{bender2020climbing}. Grounded language learning approaches respond to this concern by connecting linguistic symbols to perception, action, and interaction \parencite{bisk2020experience}.

Lexical Consensus is aligned with this grounding tradition, but it adopts a deliberately narrow operational definition. A label is considered experimentally grounded when it supports reliable image-to-label naming, label-to-image retrieval, out-of-vocabulary rejection, and resistance to falsification controls. This does not solve the full symbol grounding problem. It instead defines a measurable first layer of grounding over a frozen perceptual substrate.

The concept-carving results clarify this restricted definition. The experiments do not claim that meaning emerges from arbitrary labels alone. Instead, they show that lexical grounding depends on the relation between instruction and perceptual organization. A label becomes learnable when it stabilizes over a coherent region of experience; it becomes difficult or unstable when it is assigned to an arbitrary union of distant regions. Thus, grounding is treated not as a binary property but as a graded relation between lexical instruction and perceptual structure.

This restricted definition is important. Because DINOv2-small provides the perceptual geometry, the experiments do not demonstrate that agents acquire perception from scratch \parencite{oquab2023dinov2}. They show that artificial labels can become usable over an existing visual representation, and they measure the boundary of that usability. The frozen encoder is therefore not hidden as a confound; it is part of the experimental design. It allows us to distinguish lexical acquisition over a shared perceptual substrate from the much stronger claim that language reorganizes perception itself.

\subsection{Developmental overextension and perceptual coherence}
\label{sec:developmental_overextension}

The results are also motivated by a developmental intuition: early word
learning often begins with broad but coherent categories rather than precise
adult-like distinctions. A child may use one label for many vehicles, or one
animal sound for several related animals, before later refining the category
boundaries. \emph{Lexical Consensus} does not model child language acquisition
directly, but the concept-carving experiments operationalize a structurally
similar idea. Native categories, coherent overextensions, mid-disjunctive
concepts, and far-disjunctive concepts define a controlled gradient from
perceptually coherent naming to arbitrary set membership.

This distinction matters because it reframes the criticism that frozen encoders
already contain category structure. Human learners also begin from structured
perception. The relevant scientific question is not whether structure exists,
but how lexical instruction behaves when taught categories align with, extend,
or contradict that structure. In our experiments, coherent overextensions remain
highly learnable, while far-disjunctive concepts approach chance. This pattern
suggests that lexical acquisition over frozen perception is constrained by
perceptual coherence, not by arbitrary set membership.

The dissociation experiment in Section~\ref{sec:exp008_dissociation} strengthens
this interpretation. Because the original concept tiers were constructed using
inter-centroid distance in the same DINOv2 space used by the learners, one could
argue that the gradient simply reflects the metric used to define the tiers. The
CIFAR-100 dissociation experiment addresses this concern by separating
perceptual distance from taxonomic semantic relatedness. When the two predictors
disagree, acquisition follows perceptual distance rather than semantic distance:
pairs that are perceptually near but semantically distant remain easy to acquire,
whereas pairs that are semantically near but perceptually distant degrade. Thus,
the developmental analogy should be understood as perceptual rather than purely
semantic. What stabilizes early lexical acquisition in this framework is not
membership in an externally defined semantic class, but coherence within the
agent's perceptual substrate.

This result gives a more precise interpretation to the overextension analogy.
The framework does not claim that agents reproduce the developmental mechanisms
of children, nor that WordNet-like semantic proximity determines acquisition.
Instead, it shows that broad lexical categories are learnable when they preserve
coherence in the available perceptual geometry. In that sense, the relevant
parallel with early word learning is not the exact content of human categories,
but the structural fact that names stabilize more easily over coherent regions
of experience than over arbitrary unions of distant perceptual regions.

\subsection{Naming games, lexical conventions, and emergent communication}

The multi-agent dimension of this work is closely related to signaling games, naming games, and emergent communication. Lewis's analysis of convention framed meaning and coordination as solutions to recurring coordination problems \parencite{lewis1969convention}. In artificial agent systems, Steels and colleagues developed language-game experiments showing how shared vocabularies can emerge through interaction \parencite{steels1996spatial,steels2015talkingheads}. Naming-game models further showed how populations of agents can converge toward shared vocabularies without centralized control \parencite{baronchelli2006sharp}.

Recent emergent communication research has extended these ideas using neural agents, referential games, reinforcement learning, and differentiable communication channels \parencite{lazaridou2016multiagent,lazaridou2018emergence,havrylov2017emergence,mordatch2018emergence}. These works often study how communication protocols emerge when agents must solve cooperative tasks. Some also investigate whether the resulting protocols become compositional, interpretable, or aligned with natural language \parencite{kottur2017natural,bouchacourt2018agents}.

Lexical Consensus differs in three respects. First, the lexical forms are not invented freely by the agents; they are externally introduced as artificial labels and must become grounded through examples, retrieval, and, in the multi-agent condition, consensus. Second, the perceptual encoder is frozen, which separates visual representation from lexical stabilization. Third, the experiments are built around falsification: random labels, random embeddings, permuted bindings, OOV rejection, repeated scrambles, harder categories, no-feedback baselines, random-feedback controls, and homogeneous retrieval pools are used to determine whether the observed effect survives alternative explanations.

Thus, the framework is not primarily a study of unconstrained language emergence. It is a study of controlled lexical acquisition and stabilization. The central question is not whether agents can invent any communication protocol that solves a task, but whether a novel word can become a stable, bidirectional, and socially shared mapping over perception, and how the success of that mapping depends on perceptual coherence.

\subsection{Few-shot learning and frozen visual representations}

The single-agent setting is also related to few-shot learning, where models classify new categories from small support sets. Matching networks, prototypical networks, linear probes, and other few-shot methods provide natural baselines for evaluating whether a simple lexical learner is doing more than exploiting an embedding space \parencite{vinyals2016matching,snell2017prototypical}. Vision-language models such as CLIP also raise the question of how much semantic structure can be recovered from powerful pretrained encoders \parencite{radford2021clip}.

Our goal is not to outperform few-shot classifiers. The centroid learner is intentionally simple: it acts as a compressed prototype for a learned word. We compare it against exemplar k-NN, multi-centroid learners, logistic regression, linear SVM, and random baselines in order to locate what each mechanism contributes. The results show that linear discriminative learners can recover additional structure in difficult retrieval settings, while exemplar memory improves label-to-image retrieval over compressed prototypes. These comparisons are not used to claim classifier superiority; they are used to interpret what kinds of lexical mechanisms support naming, retrieval, and concept stabilization over frozen perception.

This distinction is important for the role of DINOv2. A frozen representation can make native-category relabeling easy, but it also imposes a measurable boundary. The concept-carving experiments show that naming accuracy follows a gradient: native categories are easiest, coherent overextensions remain learnable, and far-disjunctive concepts degrade toward chance. The contribution is therefore not the claim that a frozen encoder solves acquisition, but that a controlled lexical protocol can measure which concepts are learnable given such a substrate.

\subsection{Frozen representations and representational alignment}

A key empirical finding of this paper is that shared perceptual geometry is the dominant stabilizing force. The no-feedback baseline already achieves high consensus accuracy, and later alignment and divergence experiments show that consensus feedback does not substantially reorganize the agents' internal centroid geometry. This finding places the work in conversation with representation analysis, where researchers compare internal representations across models, layers, or training conditions \parencite{kornblith2019similarity}.

However, our use of representation comparison is intentionally simpler than methods such as CKA. We do not attempt to compare full neural activation spaces across large models. Instead, we track label centroids, inter-agent centroid distances, and alignment gains because those are the representations actually used by the lexical layer. This makes the analysis directly tied to the mechanism under study. The question is not whether two deep networks have globally similar representations, but whether lexical interaction brings agents' learned category representations closer together over time.

The answer, under the current architecture, is mostly negative. Consensus coordinates decisions over a shared visual space, but it does not act as a strong representational attractor. This result is useful because it prevents overinterpreting surface lexical agreement as evidence of deeper cognitive or perceptual change. In combination with the concept-carving results, it suggests a clearer boundary: frozen perception enables lexical acquisition for coherent regions, but lexical feedback alone does not substantially reshape the underlying perceptual geometry.

\subsection{What this framework contributes}

Taken together, prior work provides the conceptual and technical background for this study: symbol grounding motivates the need for perceptual anchoring; grounded language learning motivates bidirectional use; naming games and emergent communication show that lexical conventions can arise through interaction; few-shot learning provides baseline mechanisms; representation analysis provides tools for asking whether internal spaces align; and acquisition-based evaluation motivates the shift from static performance to learning dynamics.

The contribution of Lexical Consensus is to combine these elements into a compact empirical protocol. The framework asks a sequence of falsifiable questions: Can artificial words be grounded from few examples? Can the mapping work bidirectionally? Does the effect collapse when perception or binding is broken? Do agents converge when trained on disjoint seeds? Does consensus reshape representations, or merely coordinate decisions over an already shared geometry? And, crucially, which taught concepts become learnable as their extensions move from native categories to coherent overextensions and then to arbitrary far-disjunctive unions?

The results suggest that acquisition and consensus are achievable under controlled conditions, while stronger claims about unrestricted language acquisition and representational restructuring are not supported in the current architecture. This is the position of the paper with respect to prior work: Lexical Consensus is neither a general theory of language emergence nor a complete solution to symbol grounding. It is a reproducible experimental scaffold for studying how grounded lexical meaning can be acquired, retrieved, tested, shared, and bounded by perceptual coherence in artificial agents.


\section{Limitations and Future Scaling}
\label{sec:limitations}

The experiments reported in this paper provide a controlled empirical validation of the Lexical Consensus framework, but they should be interpreted within a constrained scope. The present study does not claim general artificial language acquisition. It operationalizes a first measurable layer of acquisition: artificial labels become usable for naming, retrieval, falsification, and, in the multi-agent condition, lexical stabilization over a structured perceptual substrate. The revised experiments extend the initial three-label setting by introducing multiple concept tiers, paired episodes, baseline learners, confidence intervals, homogeneous retrieval pools, and concept-carving evaluations. This makes the framework more robust, but still limited.

The first limitation is that the experiments remain controlled lexical episodes rather than open-ended language learning. The agents do not acquire syntax, pragmatics, dialogue strategies, compositional semantics, or embodied action. They learn word--concept mappings over visual embeddings. For this reason, the paper avoids claiming that the system learns language in the full human sense. The more precise claim is that Lexical Consensus measures how grounded lexical mappings can be acquired, retrieved, stabilized, and bounded under controlled perceptual conditions.

The second limitation is the use of a frozen perceptual encoder. DINOv2-small provides a strong and stable visual substrate \parencite{oquab2023dinov2}, but it also anchors the agents to a pretrained geometry before lexical learning begins. This limits what can be concluded about perception-changing language effects. The results do not show that agents acquire perception from scratch, nor that language reorganizes the visual encoder itself. Instead, they show which lexical concepts are learnable over a fixed perceptual substrate and where acquisition fails without representational adaptation.

The dissociation experiment (Section~\ref{sec:exp008_dissociation}) confirms
that the perceptual-coherence gradient is not an artifact of measuring within
the same space used to define concept tiers, but the limitation remains that
the encoder itself was pretrained on a large image corpus
\parencite{oquab2023dinov2}. The grounding claim applies to the lexical layer,
not to the encoder.

This limitation is also methodologically useful. A frozen encoder makes it possible to distinguish lexical acquisition over perception from perceptual learning itself. The concept-carving experiments show that the acquisition boundary is graded: native categories are easiest, coherent overextensions remain highly learnable, mid-disjunctive concepts degrade, and far-disjunctive concepts approach chance. Thus, frozen perception is not merely a confound. It is the substrate that allows the framework to measure how lexical acquisition depends on perceptual coherence.

The third limitation concerns dataset and domain breadth. The present experiments use controlled visual categories and concept tiers constructed over a limited image domain. These categories are useful because they allow transparent manipulation of perceptual coherence, inter-member distance, and native separability. However, they do not establish that the same gradient will hold across richer datasets, finer-grained domains, or more naturalistic category hierarchies. Future experiments should test larger datasets, more diverse visual domains, underrepresented categories, synthetic attribute-controlled stimuli, and fine-grained distinctions such as visually similar animal species or object subtypes.

The fourth limitation concerns concept structure. The current concept-carving experiments evaluate native categories, coherent overextensions, mid-disjunctive concepts, and far-disjunctive unions. This is sufficient to show that lexical acquisition is not arbitrary set learning, but it does not yet test compositional language. The agents do not learn phrases such as adjective--noun combinations, relational descriptions, or systematic recombinations of attributes and objects. Future work should extend the framework toward compositional vocabularies and systematic generalization, testing whether agents can acquire labels for attributes, relations, and novel combinations rather than only category-level extensions.

The fifth limitation concerns the learner architecture. The centroid learner is intentionally simple and interpretable: it represents a learned word as a compressed prototype over support examples. The experiments compare this mechanism against multi-centroid, exemplar k-NN, logistic regression, linear SVM, and random baselines. These comparisons show that different mechanisms expose different capacities: naming accuracy is primarily constrained by concept--geometry compatibility, while retrieval benefits from exemplar memory and, under hard pools, from linear discriminative boundaries. However, the present learners remain shallow mechanisms over frozen embeddings. Richer agents with trainable perception, attention over attributes, episodic memory, or neural adaptation may support different forms of acquisition.

The sixth limitation concerns bidirectional grounding. The paper shows that image-to-label naming and label-to-image retrieval are not redundant. C1 naming follows the perceptual-coherence gradient, while C2 retrieval exposes memory fidelity and candidate-set discrimination. However, the retrieval setting remains artificial: candidate pools are constructed experimentally, even when they include hard negatives and OOV contamination. Future work should test retrieval in larger and more natural candidate spaces, where the number of distractors, semantic similarity, and open-world uncertainty are closer to real use.

The seventh limitation concerns multi-agent consensus. The consensus experiments show that agents trained on disjoint seed sets can converge toward shared lexical usage, and that feedback improves agreement over a strong no-feedback baseline. However, the same experiments also show that shared perceptual geometry is the dominant stabilizing force. Consensus coordinates lexical decisions, but it does not substantially reshape the underlying representations in the current architecture. Larger populations, richer interaction protocols, alternative consensus mechanisms, and longer communication histories are needed before making broader claims about lexical stabilization in artificial agent communities. In particular, because the current consensus experiments use native concepts on which individual agents are already near the ceiling, re-running the consensus protocol over mid-disjunctive concepts, where individual accuracy leaves substantial room for feedback to act, is the most direct way to obtain an informative test of consensus effects.

The eighth limitation concerns representational restructuring. The passive alignment, mutable adapter, and regional divergence experiments consistently suggest that shared perceptual grounding dominates over lexical feedback in this architecture. This is an important negative result, but it should not be generalized beyond the tested conditions. The experiments show that lexical consensus does not substantially reshape representation here; they do not rule out stronger language-induced representational effects in systems with trainable perception, richer feedback, or developmental learning dynamics.

These limitations define the next stage of the framework. The immediate priority is not to make stronger claims, but to scale the experimental design along the dimensions revealed by the current results. Future work should test richer concept hierarchies, larger vocabularies, additional datasets, synthetic attribute spaces, compositional labels, more agents, and trainable perceptual components. In particular, a natural next step is to introduce a gradient-based lexical or perceptual adapter capable of modifying the representation itself, testing whether far-disjunctive or attribute-based concepts can become learnable when perception is allowed to adapt.

In summary, Lexical Consensus should be understood as a controlled implementation of an acquisition-based evaluation framework \parencite{vera20xxrethinking}. The present experiments show that artificial labels can be acquired, retrieved, falsified, and stabilized under constrained conditions. More importantly, they show that acquisition is graded: agents learn labels for perceptually coherent regions more reliably than labels for arbitrary unions of distant regions. This boundary is a limitation, but it is also the central empirical result of the paper. It indicates where lexical learning over frozen perception succeeds, where it fails, and what kinds of architectures are needed for the next stage of grounded language acquisition.


\section{Conclusion}
\label{sec:conclusion}

This paper introduced \emph{Lexical Consensus}, a controlled experimental framework for studying grounded lexical acquisition in artificial agents. The work was motivated by the acquisition-based evaluation program proposed in \textcite{vera20xxrethinking}, where language acquisition is presented as a more demanding and informative axis for AI evaluation than behavioral imitation alone. While the present framework does not fully implement that broader test, it provides a first empirical approximation to its central intuition: an artificial agent should be evaluated not only by what it can output, but by whether it can acquire, use, retrieve, stabilize, and share new lexical meanings from grounded experience.

The main result of the paper is that lexical acquisition over frozen perception is not arbitrary set learning. Agents do not acquire all taught concepts equally. Instead, naming accuracy follows a perceptual-coherence gradient: native visual categories are easiest to acquire, coherent overextensions remain highly learnable, mid-disjunctive concepts degrade, and far-disjunctive concepts approach chance. This pattern reframes the interpretation of grounded word learning over pretrained perception. The presence of structure in the perceptual substrate is not merely a confound; it is the condition under which acquisition becomes measurable. 
A pre-registered dissociation experiment further confirms that this gradient is not a measurement artifact. When perceptual and semantic distance are
measured independently, acquisition accuracy tracks the perceptual predictor in pairs where the two disagree. This rules out the concern that the gradient
merely reflects the metric used to define concept tiers. The relevant question is therefore not whether the substrate already contains visual organization, but which lexical concepts can be acquired given that organization.

This result also clarifies the relation between the present experiments and the first test proposed in \textcite{vera20xxrethinking}. The original motivation was not to teach isolated artificial labels as arbitrary mathematical sets, but to test whether agents can acquire usable vocabulary through grounded instruction. The concept-carving experiments bring the framework closer to that spirit. They show that broad but perceptually coherent categories are acquired more reliably than arbitrary unions of distant regions. In this sense, the framework begins to approximate a developmental pattern: early lexical acquisition stabilizes names over coherent regions of experience before finer distinctions are learned.

The bidirectional grounding results further show that naming and retrieval are not redundant. C1 evaluates whether an agent can assign the correct learned label to a new image. C2 evaluates whether a learned label can retrieve a valid perceptual instance. The experiments show that these two directions expose different capacities. Naming accuracy is primarily governed by compatibility between the taught concept and the perceptual geometry. Retrieval accuracy exposes a memory-fidelity dimension, where exemplar-based mechanisms can outperform compressed prototypes, while linear discriminative baselines recover additional structure under hard candidate pools. A complete evaluation of grounded lexical acquisition should therefore include both directions.

The falsification controls strengthen this interpretation. The grounding effect degrades or collapses when labels are randomized, embeddings are randomized, image--embedding bindings are permuted, or out-of-vocabulary examples are introduced. Repeated scrambles and homogeneous candidate-pool evaluations reduce the risk that the observed effects are artifacts of label frequency, candidate-set construction, or a single favorable split. Baseline comparisons also clarify the role of the centroid learner: it is an interpretable lexical mechanism, not a state-of-the-art classifier.

The multi-agent experiments add a complementary boundary. Agents trained on disjoint seed sets can converge toward shared lexical usage, and feedback improves agreement over a strong no-feedback baseline. However, consensus does not substantially reorganize internal representations under the current architecture. Passive centroid alignment, mutable adapter experiments, and regional divergence tests all point toward the same conclusion: shared perceptual geometry is the dominant stabilizing force. Lexical consensus coordinates decisions over perception more than it reshapes perception itself.

This distinction is important. A less careful interpretation could mistake shared label use for deep representational transformation. Our results support a more precise claim: artificial agents can acquire and stabilize lexical mappings over a common perceptual substrate, but such stabilization does not by itself imply that language has altered the underlying perceptual geometry. This is not a failure of the framework. It is one of its useful outcomes. A good acquisition-based evaluation should reveal both what a system can do and where its capacities stop.

The present work therefore captures the spirit, but not yet the full scope, of the first test proposed in \textcite{vera20xxrethinking}. We remain far from demonstrating complete artificial language acquisition, especially in the richer sense involving open-ended environments, embodied interaction, syntax, pragmatics, compositionality, and long-term conceptual development. However, we have established a methodology that makes progress in that direction possible: define novel lexical mappings, ground them in perception, test bidirectionality, vary concept coherence, introduce multi-agent stabilization, measure information transfer, compare learner mechanisms, and apply falsification controls.

Future work should scale this framework toward richer developmental settings. The most immediate extensions are larger vocabularies, additional datasets, synthetic attribute-controlled stimuli, compositional labels, finer-grained category hierarchies, more agents, richer interaction protocols, and trainable perceptual or lexical adapters. The central methodological principle should remain unchanged: claims about acquired meaning must be paired with controls showing where acquisition succeeds, where it degrades, and where it fails.

In this sense, Lexical Consensus is not presented as a final benchmark, but as a reproducible experimental scaffold. It turns a theoretical proposal about language acquisition as AI evaluation into an empirical program. If future systems are to be evaluated by their capacity to acquire and share meaning, then the path begins with experiments like this: controlled enough to be falsified, structured enough to reveal gradients of acquisition, and extensible enough to grow toward more demanding tests of artificial language understanding.

\appendix

\section{Supplementary Experimental Inventory}
\label{app:experimental-inventory}

Table~\ref{tab:appendix-experiment-inventory} summarizes the experimental sequence used in this study. The goal of the appendix is not to introduce additional claims, but to make explicit the scope of the empirical validation supporting the main results.

\begin{table*}[t]
    \centering
    \small
    \setlength{\tabcolsep}{4pt}
    \renewcommand{\arraystretch}{1.15}
    \begin{tabularx}{\textwidth}{
        >{\raggedright\arraybackslash}p{0.17\textwidth}
        >{\raggedright\arraybackslash}p{0.31\textwidth}
        >{\raggedright\arraybackslash}X
    }
        \toprule
        Experiment family & Purpose & Main outcome \\
        \midrule
        exp\_000 & Perceptual substrate diagnostic & DINOv2-small provides moderate category separability \\
        exp\_001 & Single-agent lexical acquisition & Carroll labels acquired from few examples \\
        exp\_001 C2 & Bidirectional grounding & Labels support image retrieval, not only naming \\
        exp\_002 & Grounding falsification controls & Grounding collapses when perception or binding is broken \\
        exp\_002b & Balanced-label control & Random-label artifacts unstable under repeated scrambles \\
        exp\_003 & Multi-agent consensus & Agents converge on shared lexical usage \\
        exp\_004 & Shannon and graph logging & Label assignments show high information transfer \\
        exp\_005 & Passive centroid alignment & Consensus does not substantially reshape centroid geometry \\
        exp\_005b & Active language-conditioned geometry & Adapters do not make consensus a representational attractor \\
        exp\_006 & Regional divergence & Shared perception resists divergence under tested regimes \\
        exp\_007a & Native concept baseline & Native concepts are acquired reliably across learners \\
        exp\_007b & Far-disjunctive concept stress test & Arbitrary distant unions degrade toward chance in C1 \\
        exp\_007b extension & Near/mid/far concept-carving gradient & C1 follows a perceptual-coherence gradient \\
        exp\_007b clean & Homogeneous C2 pool re-run & C2 retrieval exposes a memory-fidelity dimension \\
        exp\_008 & Perceptual vs.\ semantic dissociation & Perceptual distance governs C1; semantics adds nothing \\
        \bottomrule
    \end{tabularx}
    \caption{Summary of the experimental families included in the study.
    Several families contain multiple sub-experiments, controls, homogeneous
    re-runs, or dissociation analyses.}
    \label{tab:appendix-experiment-inventory}
\end{table*}

\section{Concept-Carving Supplement}
\label{app:concept-carving}

The concept-carving experiments extend the initial native-category acquisition
setting by testing whether artificial labels remain learnable when their
extensions depart from native visual categories. We organize taught concepts
into four tiers: native, near-disjunctive, mid-disjunctive, and
far-disjunctive. Table~\ref{tab:appendix-concept-tiers} summarizes the
interpretation of each tier.

\begin{table*}[t]
    \centering
    \small
    \setlength{\tabcolsep}{4pt}
    \renewcommand{\arraystretch}{1.15}
    \begin{tabularx}{\textwidth}{
        >{\raggedright\arraybackslash}p{0.18\textwidth}
        >{\raggedright\arraybackslash}p{0.34\textwidth}
        >{\raggedright\arraybackslash}X
    }
        \toprule
        Concept tier & Definition & Interpretation \\
        \midrule
        Native
        & One label corresponds to one native category
        & Direct naming of a coherent visual region \\
        
        Near-disjunctive
        & One label groups nearby or related categories
        & Coherent overextension, analogous to broad early categories \\
        
        Mid-disjunctive
        & One label groups moderately distant categories
        & Partial conflict with perceptual geometry \\
        
        Far-disjunctive
        & One label groups distant categories
        & Arbitrary union across distant perceptual regions \\
        \bottomrule
    \end{tabularx}
    \caption{Concept tiers used in the concept-carving experiments. The tiers
    are organized by the relationship between the taught lexical concept and
    the frozen perceptual geometry.}
    \label{tab:appendix-concept-tiers}
\end{table*}

Table~\ref{tab:c1-gradient} in the main text reports the representative C1
naming results for the homogeneous \(3\)-way, \(5\)-shot setting. These values
support the claim that lexical acquisition over frozen perception is constrained
by perceptual coherence: native categories are acquired most reliably,
coherent overextensions remain highly learnable, mid-disjunctive concepts
degrade, and far-disjunctive concepts approach chance.

The purpose of this appendix is therefore not to duplicate the numerical C1
gradient table, but to make explicit how the concept tiers used in that
evaluation should be interpreted.

\section{C2 Retrieval Pool Conditions}
\label{app:c2-pools}

The C2 retrieval experiments evaluate whether a learned label can recover a
valid image from a candidate pool. Because retrieval accuracy is sensitive to
candidate-set construction, the final C2 analysis uses homogeneous pool
conditions across all concept tiers. Table~\ref{tab:appendix-c2-pools}
summarizes the pool types.

\begin{table}[t]
    \centering
    \small
    \begin{tabular}{ll}
        \toprule
        Pool type & Description \\
        \midrule
        Standard N-way & Candidates sampled from the concepts present in the episode \\
        Small & Five candidates \\
        Medium & Ten candidates \\
        Large & Twenty-five candidates \\
        Hard & Nearest-neighbor distractors from non-target concepts \\
        OOV-25 & Candidate pool with approximately 25\% out-of-vocabulary distractors \\
        OOV-50 & Candidate pool with approximately 50\% out-of-vocabulary distractors \\
        \bottomrule
    \end{tabular}
    \caption{Candidate-pool constructions used for homogeneous C2 retrieval evaluation.}
    \label{tab:appendix-c2-pools}
\end{table}

Table~\ref{tab:c2-hard-pool} in the main text reports representative hard-pool
C2 retrieval results. Hard pools are particularly informative because
distractors are visually close to the target concept in the frozen embedding
space. The main result is that exemplar memory improves over compressed
centroid prototypes across all concept tiers, while logistic regression
recovers additional discriminative structure from the frozen embedding space.

The corresponding retrieval accuracies under each pool construction are
reported in Figure~\ref{fig:exp007-c2-pool-sensitivity} of the main text.

The purpose of this appendix is not to duplicate the hard-pool numerical
results reported in Table~\ref{tab:c2-hard-pool}, nor the pool-sensitivity
figure reported in Figure~\ref{fig:exp007-c2-pool-sensitivity}, but to document
the candidate-pool constructions used to evaluate the robustness of C2
retrieval.

\section{NMI Gate Diagnostic Audit}
\label{app:nmi-audit}

During the concept-carving experiments, an NMI-based pre-episode gate was initially explored as a diagnostic for whether a candidate disjunctive concept behaved like a native category. The audit showed that this quantity was non-discriminative for balanced two-category unions.

The pre-episode diagnostic computed normalized mutual information between a binary membership variable and the ten-way native category label. The binary variable indicates whether an image belongs to a two-category union:

\begin{equation}
    B(x) =
    \begin{cases}
    1, & y(x) \in \{y_u,y_v\}, \\
    0, & \text{otherwise}.
    \end{cases}
\end{equation}

For a balanced dataset with ten categories and two categories in the union,

\begin{equation}
    P(B=1)=0.20,
    \qquad
    P(B=0)=0.80.
\end{equation}

The entropy of the binary membership variable is therefore:

\begin{equation}
    H(B)
    =
    -0.20\log_2(0.20)
    -
    0.80\log_2(0.80)
    =
    0.7219.
\end{equation}

Since the native category determines whether an image belongs to the two-category union,

\begin{equation}
    I(B;Y) = H(B).
\end{equation}

With ten balanced native categories,

\begin{equation}
    H(Y)=\log_2(10)=3.3219.
\end{equation}

Using the average-normalized form,

\begin{equation}
    \operatorname{NMI gate}(B,Y)
    =
    \frac{2I(B;Y)}{H(B)+H(Y)}
    =
    \frac{2 \cdot 0.7219}{0.7219 + 3.3219}
    =
    0.3570.
\end{equation}

This value depends only on the balanced two-from-ten construction. It is therefore constant across all two-category unions and provides no filtering power for pair selection. For this reason, the NMI gate is not used as evidence for concept selection in the paper. The final concept tiers are organized using inter-member distance and native separability diagnostics.

\section{Supplementary Figures}
\label{app:supplementary-figures}

The main text reports the figures needed to support the core argument. Additional diagnostic and control figures are provided here for transparency.

\begin{figure}[t]
    \centering
    \includegraphics[width=0.85\linewidth]{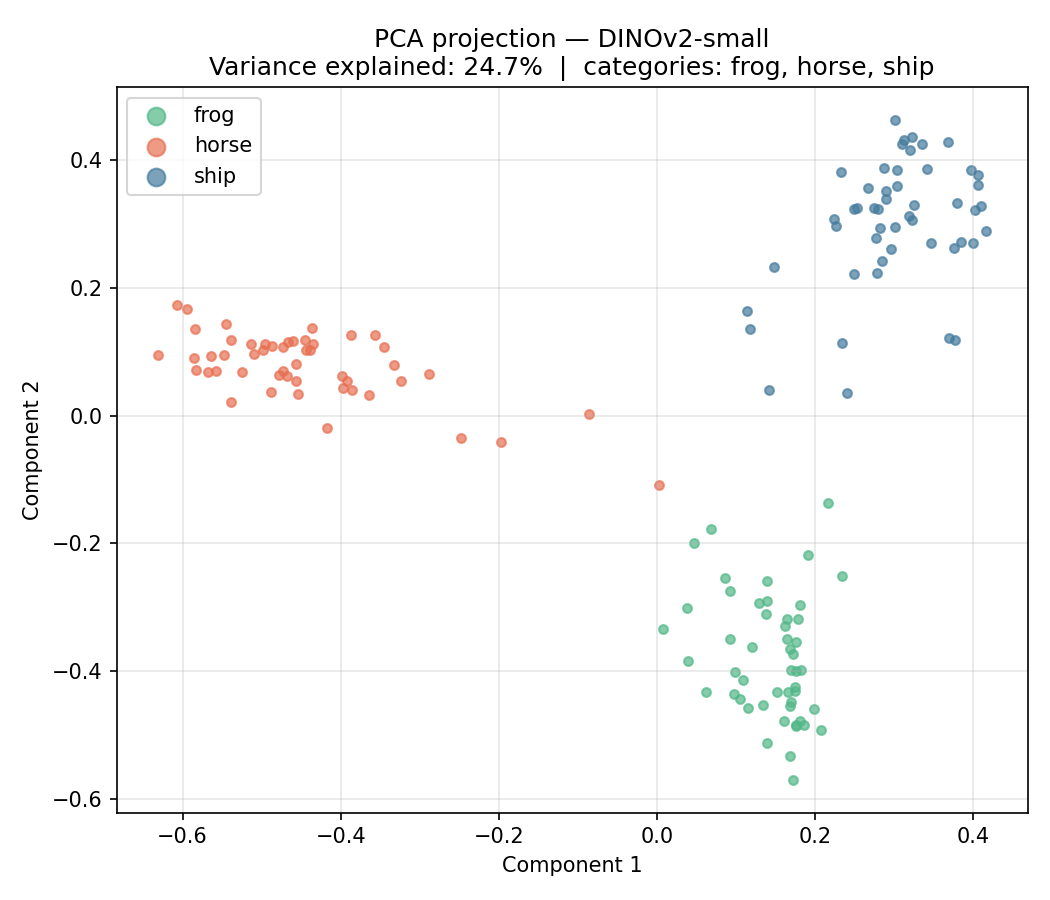}
    \caption{PCA projection of the frozen DINOv2-small embedding space for the initial visual categories.}
    \label{fig:app-exp000-pca}
\end{figure}

\begin{figure*}[t]
    \centering
    \begin{subfigure}{0.32\textwidth}
        \centering
        \includegraphics[width=\linewidth]{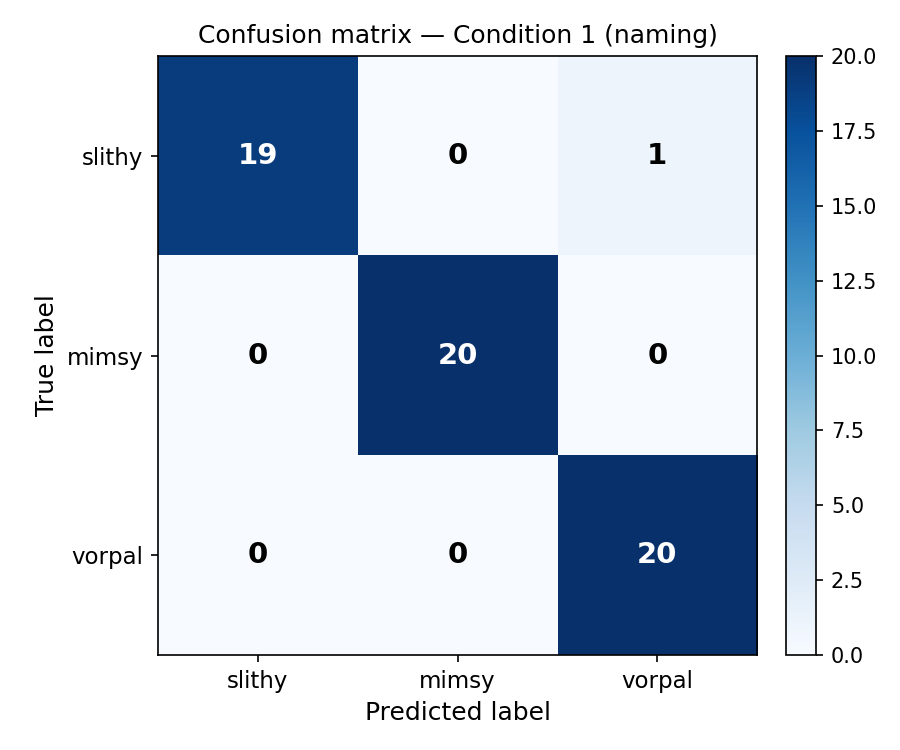}
        \caption{5 seeds}
    \end{subfigure}
    \hfill
    \begin{subfigure}{0.32\textwidth}
        \centering
        \includegraphics[width=\linewidth]{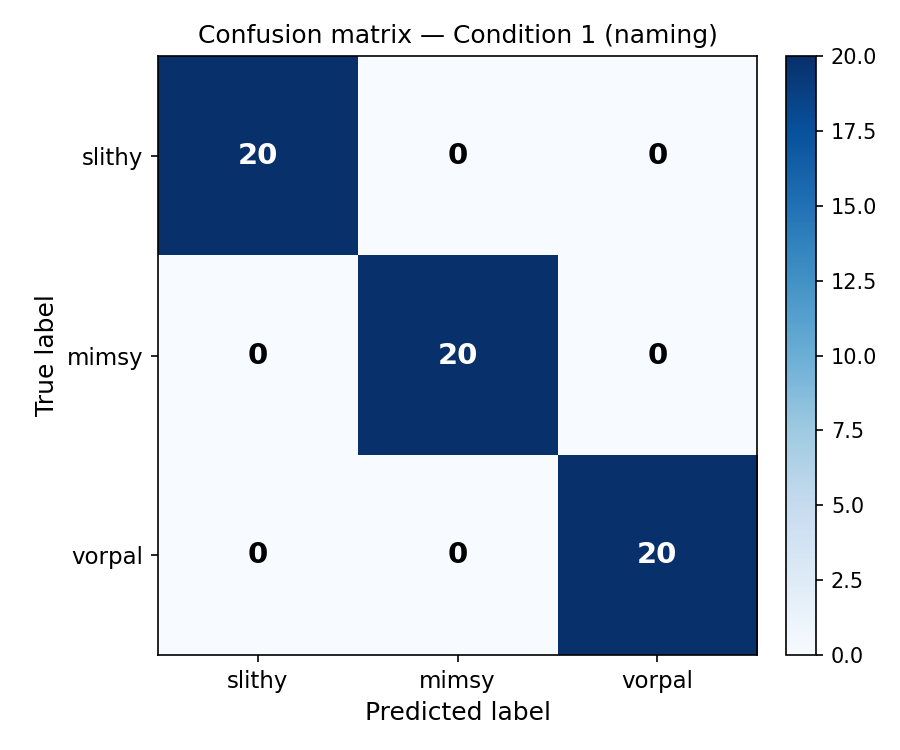}
        \caption{10 seeds}
    \end{subfigure}
    \hfill
    \begin{subfigure}{0.32\textwidth}
        \centering
        \includegraphics[width=\linewidth]{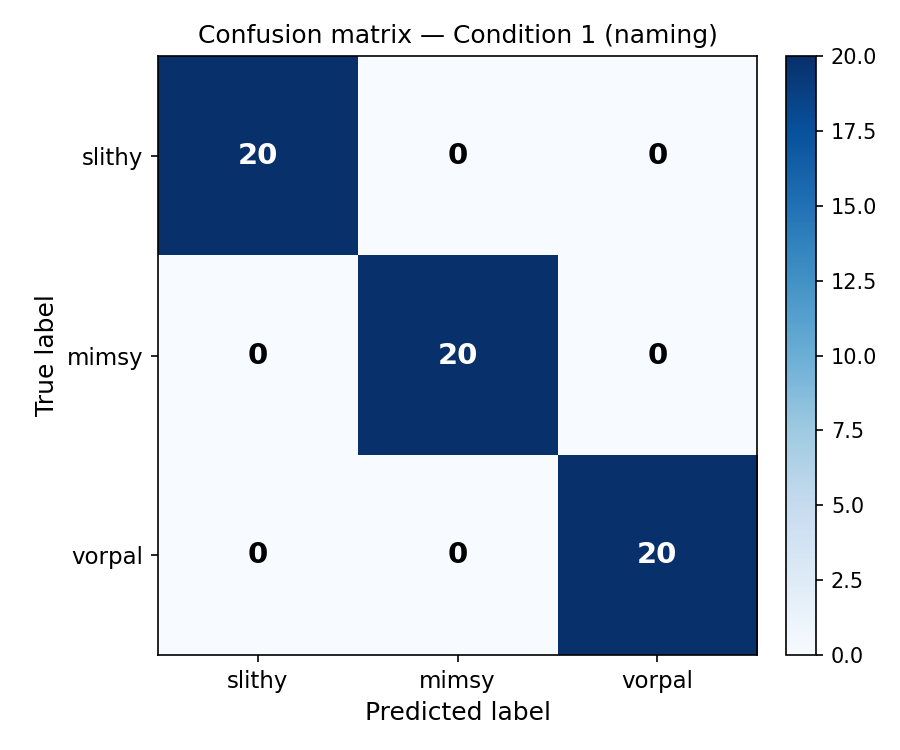}
        \caption{15 seeds}
    \end{subfigure}
    \caption{Single-agent confusion matrices across seed sizes. The mapping saturates after 10 examples per label in the initial native-category setting.}
    \label{fig:app-exp001-confusions}
\end{figure*}

\begin{figure*}[t]
    \centering
    \begin{subfigure}{0.32\textwidth}
        \centering
        \includegraphics[width=\linewidth]{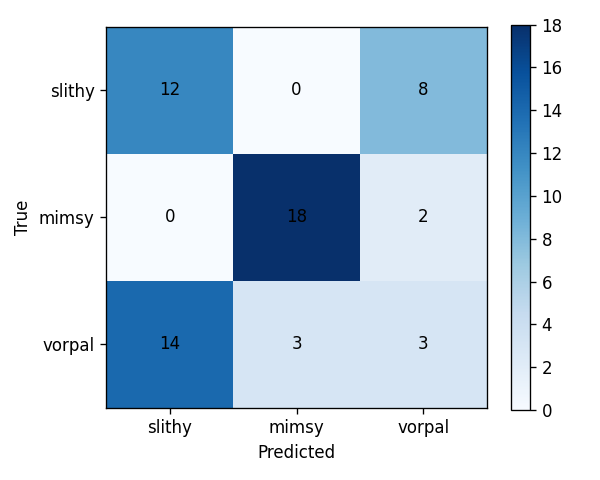}
        \caption{Random labels}
    \end{subfigure}
    \hfill
    \begin{subfigure}{0.32\textwidth}
        \centering
        \includegraphics[width=\linewidth]{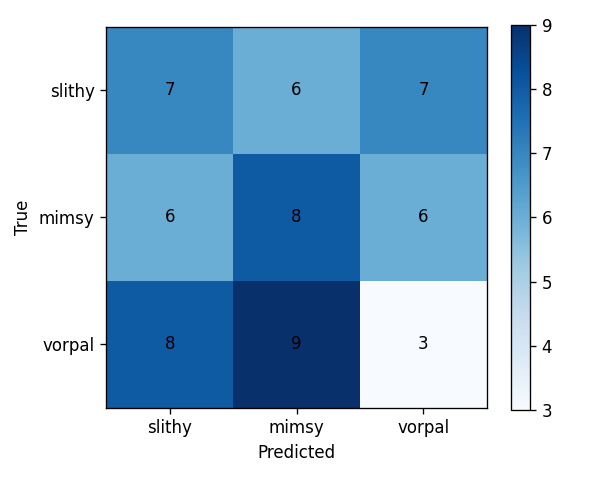}
        \caption{Random embeddings}
    \end{subfigure}
    \hfill
    \begin{subfigure}{0.32\textwidth}
        \centering
        \includegraphics[width=\linewidth]{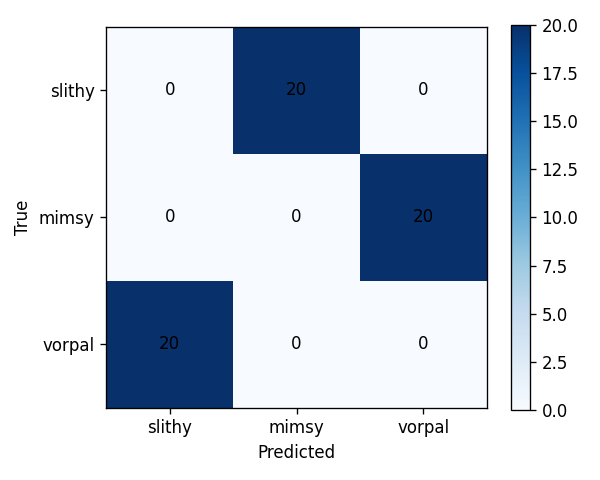}
        \caption{Permuted embeddings}
    \end{subfigure}
    \caption{Grounding-control confusion matrices for C1. Performance degrades or collapses when labels, embeddings, or image--embedding bindings are disrupted.}
    \label{fig:app-exp002-controls}
\end{figure*}

\begin{figure}[t]
    \centering
    \includegraphics[width=0.85\linewidth]{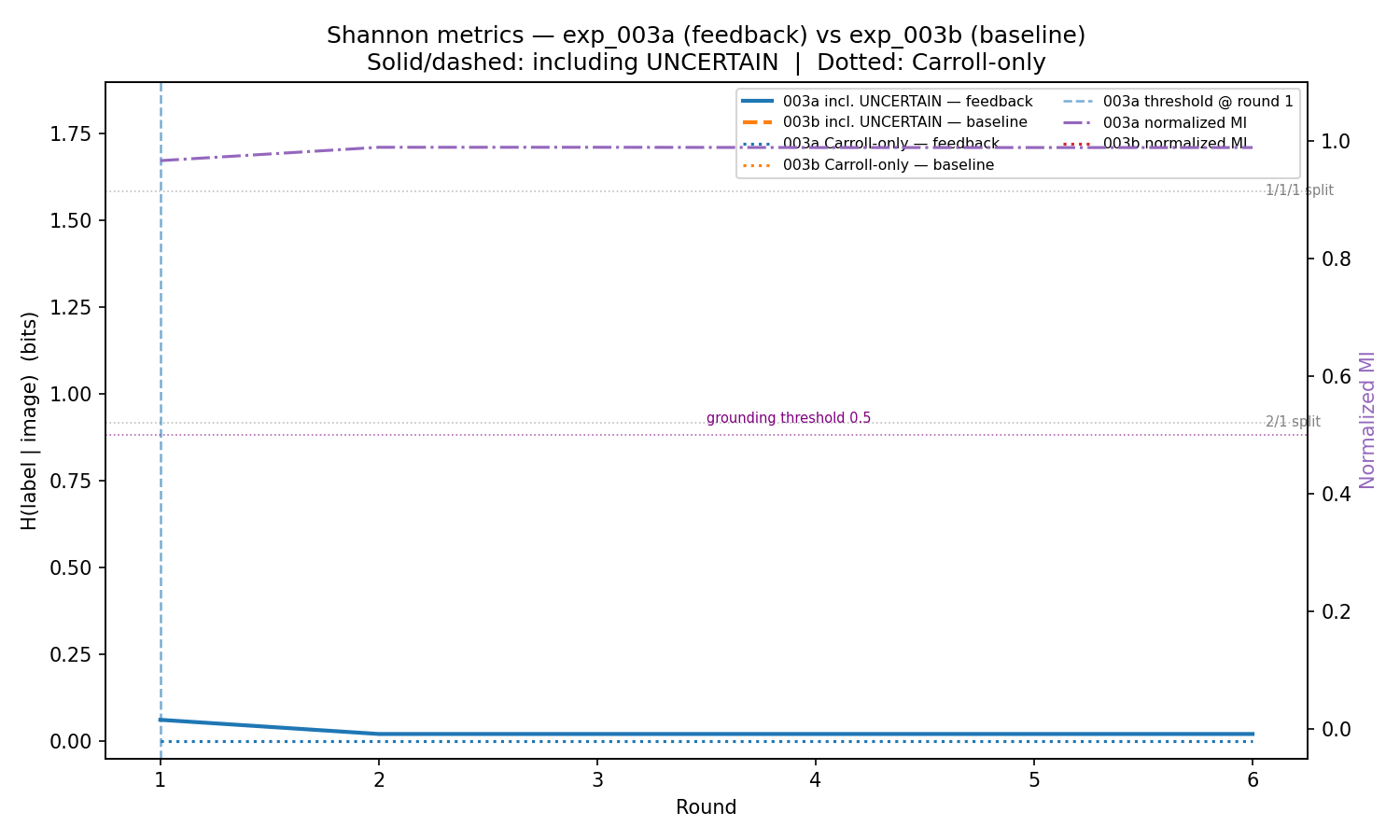}
    \caption{Entropy curve from the information-theoretic analysis. The curve provides an additional view of lexical stabilization over rounds.}
    \label{fig:app-exp004-entropy}
\end{figure}

\begin{figure}[t]
    \centering
    \includegraphics[width=0.85\linewidth]{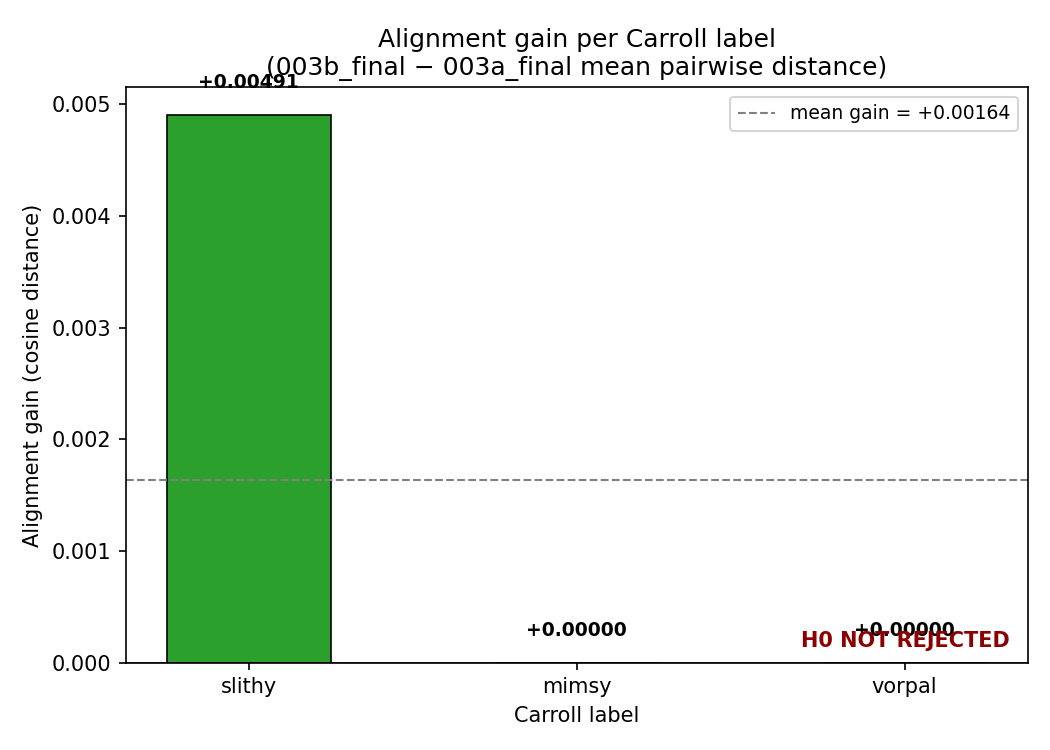}
    \caption{Alignment gain by label in the passive centroid-alignment experiment. Gains remain small, supporting the null result reported in the main text.}
    \label{fig:app-exp005-gain}
\end{figure}

\begin{figure}[t]
    \centering
    \includegraphics[width=0.85\linewidth]{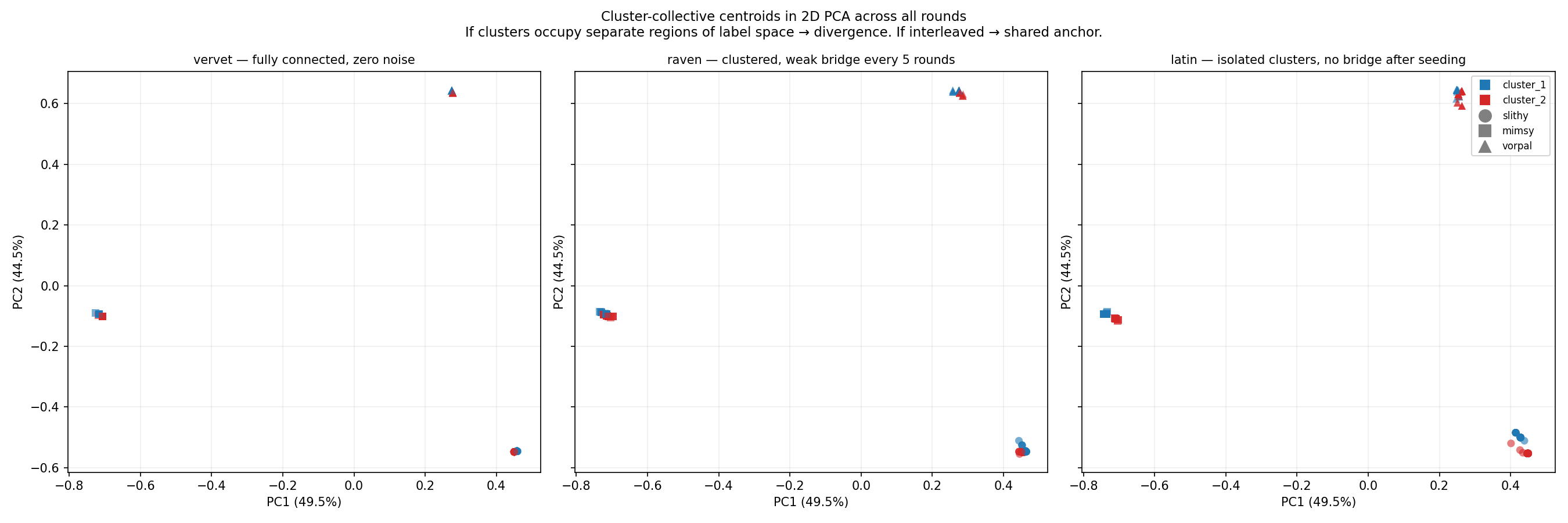}
    \caption{PCA visualization of cluster centroids in the regional-divergence experiment.}
    \label{fig:app-exp006-cluster-pca}
\end{figure}

\section{Implementation Parameters}
\label{app:implementation-parameters}

The experiments use a frozen DINOv2-small encoder as the perceptual substrate. The encoder produces 384-dimensional embeddings from the CLS token, which are normalized before cosine-similarity computations. Lexical centroids are computed as normalized means of the embeddings assigned to each artificial label.

The initial native-category experiments use a small Carroll lexicon applied to categories drawn from the CIFAR-10 benchmark \parencite{krizhevsky2009learning}, with controlled seed sets of $5$, $10$, and $15$ examples per label. The concept-carving experiments extend to all ten CIFAR-10 categories and use homogeneous $3$-way, $5$-shot episodes with $30$ paired episodes. All images are encoded once using frozen DINOv2-small embeddings and cached; no data augmentation is applied. Learners are evaluated on identical support and query splits in each paired episode. The learner roster includes centroid, multi-centroid, exemplar k-NN, logistic regression, linear SVM, and random baselines.

The multi-agent consensus mechanism uses a convergence threshold of $\tau=0.70$. A consensus label is accepted when at least 70\% of the agent population assigns the same label to a given item. Out-of-vocabulary rejection is computed from the maximum association score between an input embedding and the learned lexical labels. In the reported OOV control, the optimal threshold was $\gamma=0.5289$.

All reported information-theoretic quantities are computed from empirical label distributions. Conditional entropy is estimated from the per-image distribution of agent assignments, and normalized mutual information is computed as

\begin{equation}
    \operatorname{NMI}(X,L) = \frac{I(X;L)}{H(L)}.
\end{equation}

When $H(L)=0$, normalized mutual information is defined as $0$, since this corresponds to lexical collapse rather than successful grounding. This NMI is distinct from the pre-episode concept-gate diagnostic audited in Appendix~\ref{app:nmi-audit}.

\section{Dissociation Experiment Regression Supplement}
\label{app:exp008_regression}

\begin{figure}[t]
    \centering
    \includegraphics[width=\linewidth]{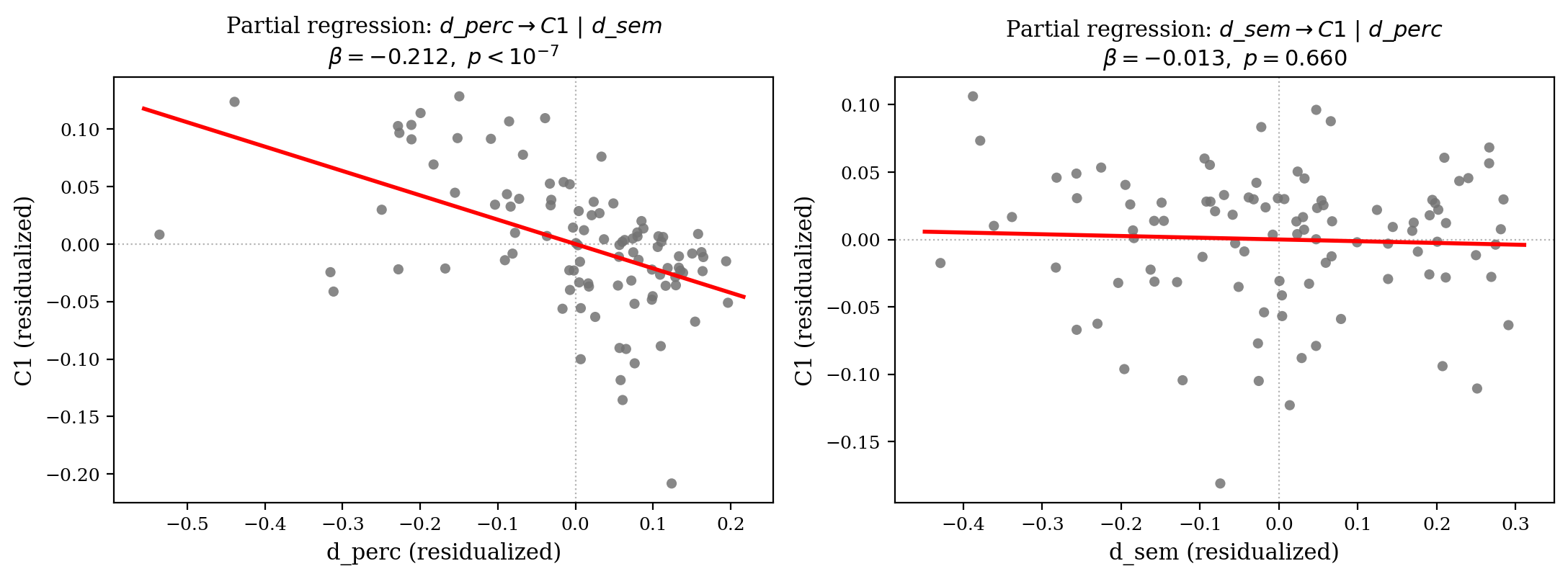}
    \caption{Partial-regression diagnostic for the CIFAR-100 dissociation
    experiment. After controlling for semantic distance, perceptual distance
    remains a strong predictor of C1 naming accuracy. After controlling for
    perceptual distance, semantic distance contributes no significant
    explanatory power.}
    \label{fig:exp008_partial_regression}
\end{figure}

Figure~\ref{fig:exp008_partial_regression} shows the partial-regression
diagnostic for the CIFAR-100 dissociation experiment. Table~\ref{tab:exp008_regression}
reports the corresponding regression and model-comparison statistics.

\begin{table}[t]
\centering
\begin{tabular}{lccc}
\toprule
Predictor or comparison & Estimate & Statistic & \(p\)-value \\
\midrule
Perceptual distance & \(\beta_{\mathrm{perc}} = -0.212\) & partial \(R^2 = 0.245\) & \(< 10^{-7}\) \\
Semantic distance & \(\beta_{\mathrm{sem}} = -0.013\) & partial \(R^2 = 0.002\) & \(0.660\) \\
Semantic added to perceptual model & -- & \(\chi^2 = 0.20\) & \(0.655\) \\
Perceptual added to semantic model & -- & \(\chi^2 = 28.14\) & \(< 10^{-7}\) \\
CIFAR-100 superclass match & \(\beta = 0.015\) & -- & \(0.528\) \\
\bottomrule
\end{tabular}
\caption{Regression and robustness statistics for the CIFAR-100 dissociation
experiment. Perceptual distance explains C1 naming accuracy, while semantic
distance and superclass membership add no significant explanatory power after
controlling for perceptual distance.}
\label{tab:exp008_regression}
\end{table}

\section{Reproducibility Statement}
\label{app:reproducibility}

The implementation logs experimental configurations, random seeds, image identifiers, concept assignments, learner types, support/query splits, candidate-pool construction, agent assignments, consensus events, and per-round metrics. The results are stored as structured JSON, CSV, and image artifacts. The experiment graph is additionally stored in Neo4j for selected runs, allowing the acquisition and consensus process to be inspected as a temporal graph.

The code and experiment outputs are organized so that each experimental family can be rerun independently. The concept-carving experiments additionally use paired episodes, ensuring that all learners are evaluated on identical support and query splits. This supports paired statistical testing and reduces variance from split differences.

The intended role of this appendix is to make the empirical basis of the paper inspectable without expanding the claims beyond the controlled setting reported in the main text. 
Code, experiment ledgers, figure-generation scripts, and the internal pre-registration record are publicly available in the \href{https://github.com/patriciomvera/lexical-consensus}{project repository}.

\section{Pre-registration Record for the Dissociation Experiment}
\label{app:exp008-preregistration}

The dissociation experiment reported in Section~\ref{sec:exp008_dissociation} was governed by decision rules fixed before any Phase 3 episode data were collected. The full specification, including the rules reproduced below, is included in the project repository as \texttt{experiments/exp\_008\_dissociation/exp\_008\_preregistration.md}, dated 2026-06-08. We use the term \emph{pre-registered} in an internal sense: the decision rules were written before the dissociation episode analysis and are reproduced here for transparency; no external registry was used

Rule A, corresponding to ``perception governs; circularity closed,'' required
all of the following criteria:
\begin{enumerate}
    \item[(A1)] In the two-predictor model, \(\beta_{\mathrm{perc}}\) is
    significant (\(p < 0.05\)) and negative.
    
    \item[(A2)] \(\beta_{\mathrm{sem}}\) is not significant
    (\(p \geq 0.10\)) after controlling for perceptual distance, or is
    significant with partial \(R^2\) less than one third of the perceptual
    partial \(R^2\).
    
    \item[(A3)] In the dissociation subset (Q3+Q4),
    \(\mathrm{Spearman}(\mathrm{C1}, d_{\mathrm{perc}})\) is negative with
    \(|\rho| > 0.30\) and exceeds
    \(|\mathrm{Spearman}(\mathrm{C1}, d_{\mathrm{sem}})|\).
    
    \item[(A4)] The median C1 accuracy of Q4, corresponding to perceptually
    near but semantically far pairs, is greater than the median C1 accuracy of
    Q3, corresponding to perceptually far but semantically near pairs.
\end{enumerate}

Rule B, corresponding to ambiguous dissociation, triggered on any of the
following conditions: comparable partial \(R^2\) for both predictors; weak
correlations or insufficient pairs in the dissociation subset; or
\(\mathrm{VIF} > 5\), indicating severe collinearity.

Rule C, corresponding to ``semantics wins or gradient breaks,'' triggered on
any of the following conditions: \(\beta_{\mathrm{perc}}\) is non-significant
while \(\beta_{\mathrm{sem}}\) is significant;
\(|\rho_{\mathrm{sem}}|\) exceeds \(|\rho_{\mathrm{perc}}|\) by more than
\(0.15\) in the dissociation subset; or the median C1 accuracy of Q3 exceeds
that of Q4.

The observed results satisfied all four criteria of Rule A:
\(\beta_{\mathrm{perc}} = -0.212\), \(p < 10^{-7}\) (A1);
\(\beta_{\mathrm{sem}}\), \(p = 0.660\) (A2);
\(\rho_{\mathrm{perc}} = -0.475\) versus
\(\rho_{\mathrm{sem}} = +0.327\) in Q3+Q4 (A3); and median Q4
\(= 0.876\) greater than median Q3 \(= 0.837\) (A4).

One documented post-hoc correction affects the decision-rule logic and is
reported in full. The initial analysis computed the variance inflation factor
without the intercept column, a known \texttt{statsmodels} artifact, yielding
\(\mathrm{VIF} = 9.69\). Taken at face value, this nominally satisfied the
collinearity disjunct of Rule B, creating a conflict between rules for which
the pre-registration specified no precedence. The error was detected through
internal consistency checking: a VIF of \(9.69\) is mathematically incompatible
with the directly measured Pearson correlation between the two predictors on
the sampled pairs (\(r = 0.102\), which entails
\(\mathrm{VIF} = 1/(1-r^2) = 1.01\)) and with the Phase 1 diagnostic over all
\(4{,}950\) pairs (\(r = 0.155\)). The corrected computation yields
\(\mathrm{VIF} = 1.01\). This correction changes no regression coefficients,
\(p\)-values, \(R^2\), or partial \(R^2\) values; its sole effect is to remove
the spurious Rule B trigger, so that Rule A holds cleanly and without
ambiguity. We report this deviation in full because pre-registration is only
meaningful when departures from the registered analysis are documented rather
than silently resolved.

One minor deviation from the registered specification is also noted: the
backward-compatibility sanity check of the original exp\_007 tier pairs
against the CIFAR-100 distance matrices (specification Section~5.3) was not
included in the run report.

\printbibliography
\end{document}